\documentclass[runningheads]{llncs}

\usepackage{eccv}

\usepackage{eccvabbrv}

\usepackage{graphicx}
\usepackage{booktabs}

\usepackage{makecell}
\usepackage{multirow}
\usepackage{algorithm}
\usepackage{algorithmic}
\usepackage{enumitem}
\setlist[itemize]{topsep=0pt}

\newcommand{\redContent}[1]{{\color{red}#1}}
\newcommand{\blueContent}[1]{{\color{blue}#1}}

\newcommand{\argminE}{\mathop{\mathrm{argmin}}}   

\usepackage[accsupp]{axessibility}  

\usepackage[breaklinks,colorlinks]{hyperref}

\usepackage{orcidlink}

\begin{document}



\title{Source Prompt Disentangled Inversion for Boosting Image Editability with  Diffusion Models}


\author{Ruibin Li\inst{1} \and
Ruihuang Li\inst{1} \and
Song Guo\inst{2} \and
Lei Zhang\inst{1}\thanks{Corresponding author. }}

\authorrunning{R. Li et al.}
\titlerunning{SPDInv for Boosting Image Editablity with Diffusion Models}

\institute{The Hong Kong Polytechnic University \and
The Hong Kong University of Science and Technology
}


\maketitle

\begin{abstract}
Text-driven diffusion models have significantly advanced the image editing performance by using text prompts as inputs. 
One crucial step in text-driven image editing is to invert the original image into a latent noise code conditioned on the source prompt. 
While previous methods have achieved promising results by refactoring the image synthesizing process, the inverted latent noise code is tightly coupled with the source prompt, limiting the image editability by target text prompts.
To address this issue, we propose a novel method called \textbf{S}ource \textbf{P}rompt \textbf{D}isentangled \textbf{Inv}ersion (SPDInv), which aims at reducing the impact of source prompt, thereby enhancing the text-driven image editing performance by employing diffusion models.
To make the inverted noise code be independent of the given source prompt as much as possible, we indicate that the iterative inversion process should satisfy a fixed-point constraint. Consequently, we transform the inversion problem into a searching problem to find the fixed-point solution, and utilize the pre-trained diffusion models to facilitate the searching process.
The experimental results show that our proposed SPDInv method can effectively mitigate the conflicts between the target editing prompt and the source prompt, leading to a significant decrease in editing artifacts.
In addition to text-driven image editing, with SPDInv we can easily adapt customized image generation models to localized editing tasks and produce promising performance. The source code are available at \href{https://github.com/leeruibin/SPDInv}{https://github.com/leeruibin/SPDInv}.

\end{abstract}


\section{Introduction}
\label{sec:intro}

The emergence of diffusion models~\cite{DDPM,song2020score}, especially the Latent Diffusion Models (LDMs)\cite{LDM}, has revolutionized the field of image generation. Leveraging the exceptional semantic understanding ability of pre-trained LDMs, researchers have successfully applied them to numerous downstream tasks, such as text-to-image\cite{zhang2023adding,ramesh2022hierarchical,podell2023sdxl,peebles2023scalable}, style transfer\cite{zhang2023inversion,wang2023stylediffusion,yang2023paint}, text-to-video\cite{blattmann2023align,guo2023animatediff,chai2023stablevideo}, text-to-3D\cite{liu2023syncdreamer,liu2023zero,poole2022dreamfusion}, as well as text-driven image editing \cite{kawar2023imagic,brooks2023instructpix2pix,avrahami2022blended,nichol2021glide}. It has been demonstrated \cite{Prompt2Prompt,masactrl,plugandplay} that by delicately controlling the attention layer in LDMs, we can achieve complex image editing by modifying only the text prompts.

\begin{figure}[tb]
  \centering
  \includegraphics[width=0.9\linewidth]{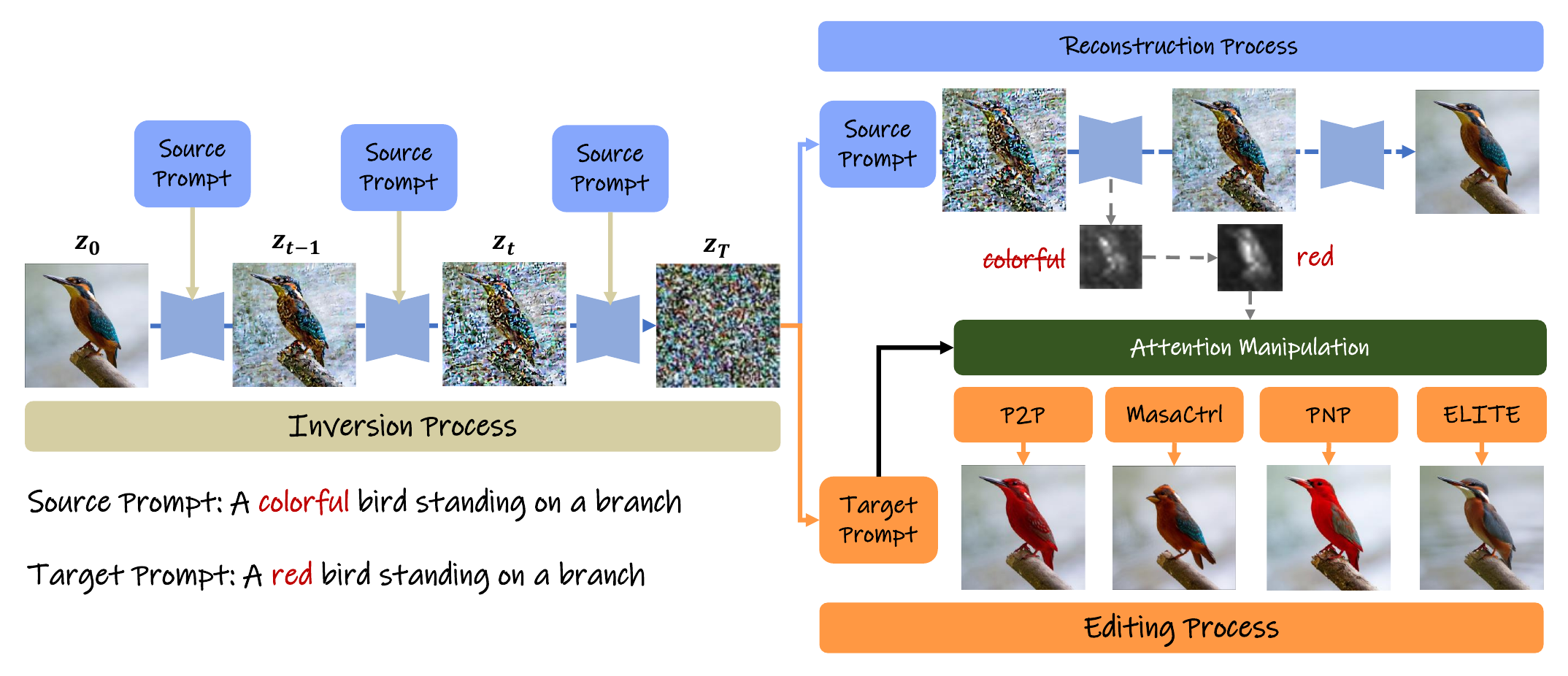}
    \vspace{-3mm}
  \caption{ Illustration of text-driven image editing pipeline.}
  \label{fig:edit_process}
  \vspace{-6mm}
\end{figure}

\begin{figure}[tb]
  \centering
  \includegraphics[width=\linewidth]{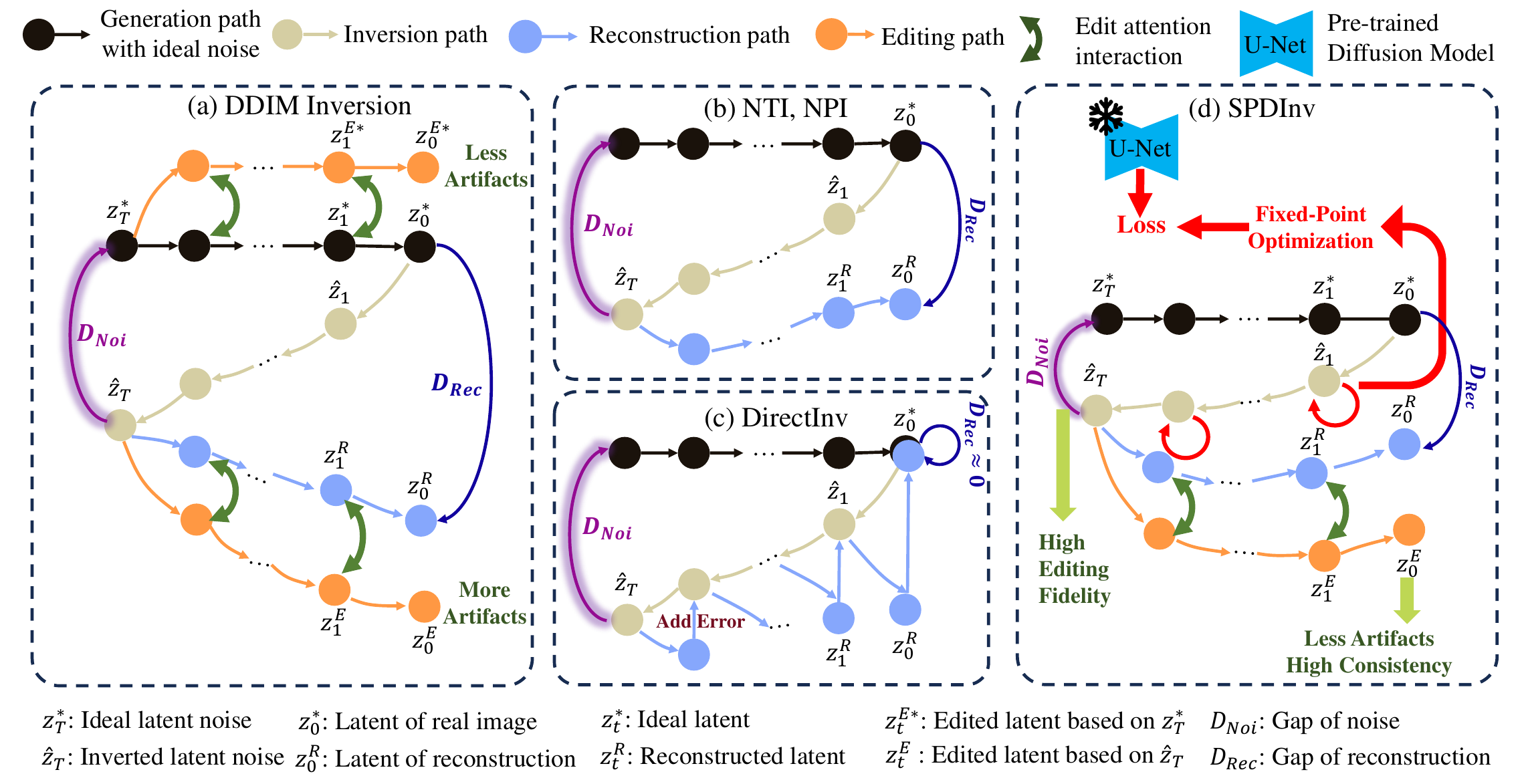}
\vspace{-5mm}
  \caption{Pipelines of different inversion methods in text-driven editing. (a) DDIM inversion inverts a real image to a latent noise code, but the inverted noise code often results in large gap of reconstruction $D_{Rec}$ with higher CFG parameters. (b) NTI optimizes the null-text embedding to narrow the gap of reconstruction $D_{Rec}$, while NPI further optimizes the speed of NTI. (c) DirectInv records the differences between the inversion feature and the reconstruction feature, and merges them back to achieve high-quality reconstruction. (d) Our SPDInv aims to minimize the gap of noise $D_{Noi}$, instead of $D_{Rec}$, which can reduce the impact of source prompt on the editing process and thus reduce the artifacts and inconsistent details encountered by the previous methods.}
  \label{fig:inversion_process}
  \vspace{-3mm}
\end{figure}

Originally, the scope of text-driven image editing was limited to images generated by the LDM. However, it was soon found that we can first convert the real image into latent noise through slight modification of the DDIM\cite{DDIM} inference pipeline along with the source prompt so that high-quality reconstruction of the original image can be obtained. 
The inverted latent noise can then be used to perform sophisticated image editing by interacting with the target prompts, as illustrated in Fig. \ref{fig:edit_process}. The inversion process (please refer to Fig. \ref{fig:inversion_process}(a)), which is crucial for achieving real-image editing, significantly impacts the editing results. Since the advanced LDMs are mostly driven by the Classifier-Free Guidance (CFG) technology\cite{CFG}, reconstruction failures often occur when the CFG parameter needs to be set higher. To address this issue, Mokady \etal~\cite{NTI} proposed the Null-Text Inversion (NTI) to optimize the null-text embedding during the reconstruction process without cumbersome tuning of model weights. The Negative-Prompt Inversion (NPI) was further developed to reduce the optimization time\cite{NPI,Prox}. Both NTI and NPI narrow the gap of reconstruction in inversion-based editing, as depicted in Fig. \ref{fig:inversion_process}(b). Ju \etal~\cite{directinv} challenged the optimization-based inversion methods and proposed Direct Inversion (DirectInv) by recording the differences between the inverted and the reconstructed features. The differences are then merged into the inference process to ensure high-quality reconstruction, as shown in Fig. \ref{fig:inversion_process}(c).

In addition to NTI, NPI and DirectInv, there are many other image inversion methods \cite{tsaban2023ledits,huberman2023edit,Prox,cho2024noise,epstein2024diffusion,zhang2024real,xu2023inversion} that have been proposed in recent years. While they can successfully reconstruct the source image by refactoring the image synthesis process, they all rely on DDIM inversion to provide the latent noise code. However, DDIM inversion assumes that the Ordinary Differential Equation (ODE) derived from the DDIM sampling process can be reversed with infinitesimally small steps. This assumption and the corresponding inversion equation exhibit instability during the inversion process. Consequently, conditioned on the given source prompt, the inverted latent noise code $\hat{z}_T$ of the source image will be closely coupled with the source prompt (please refer to \cref{sec:preliminary} for our detail analysis), leading to a significant divergence $D_{noi}$  with the ideal latent noise code $z_T^*$ (as depicted in Fig. \ref{fig:inversion_process}(a)), which is supposed to be independent to the source prompt. The dependency on the source prompt brings obstacles for editing $\hat{z}_T$ with the target prompt, resulting in artifacts and inconsistent details in the edited image. 

To address the above problem and disentangle the inverted latent noise code $\hat{z}_T$ from the source prompt, we revisit the DDIM sampling process, which iteratively denoises latent feature $z_t$ to $z_{t-1}$ with latent noise $z_T^*$ as the starting point. By reversing the order of $z_{t-1}$ and $z_t$ in the ODE formula of DDIM, we can derive the inversion formula to obtain the ideal latent noise $z_T^*$, which does not have an analytical solution but can be solved as a fixed-point problem, as discussed in the work of accelerated iterative diffusion inversion (AIDI)~\cite{AIDI}. This finding implies that the inversion process should adhere to a fixed-point constraint in order to disentangle the inverted noise code from the given source prompt. Unfortunately, as shown in Fig. \ref{fig:inversion_process}(a-c), previous efforts \cite{DDIM,NTI,NPI,directinv} have mostly focused on designing elaborate manipulations to reduce the gap of reconstruction, denoted by $D_{Rec}$, in the synthesizing process, without considering the adverse impact of the source prompt on the inverted noise code and how to reduce the gap of noise, denoted by $D_{Noi}$.

Based on the above analysis, we propose a \textbf{S}ource \textbf{P}rompt \textbf{D}isentangled \textbf{Inv}ersion method, termed as \textbf{SPDInv}. As illustrated in Fig. \ref{fig:inversion_process}(d), SPDInv aims to minimize  $D_{Noi}$, instead of $D_{Rec}$, so that the inverted noise code $\hat{z}_T$ can be independent to the source prompt as much as possible, reducing its potential conflicts with the target prompt during the editing process. To achieve this goal, we transform each inversion step into a search problem with a fixed-point constraint. 
Different from AIDI\cite{AIDI} which searches the solution by direct iteration, we reformulate the fixed-point constraint to a loss function and leverage the powerful pre-trained diffusion models to perform the searching, largely narrowing the gap between the inverted noise code and the ideal noise without any source prompt prior. Our proposed SPDInv can be easily integrated into the inversion-based text-driven editing pipelines, such as P2P~\cite{Prompt2Prompt}, MasaCtrl~\cite{masactrl} and PNP~\cite{plugandplay}, with just 10 lines of codes, significantly alleviating the artifacts and inconsistencies in the edited images. Furthermore, SPDInv can be easily applied to those customized text-to-image generation methods such as ELITE~\cite{wei2023elite}, adapting them to text-driven localized editing tasks with minor modifications. 
The main contributions of this paper are summarized as follows:

\begin{itemize}
\item We present SPDInv, a plug-and-play inversion method designed for text-driven image editing. It harnesses the power of pre-trained diffusion models to perform fixed-point searching in the inversion process, disentangling the inverted noise code from the source prompt as much as possible.
\item We show that SPDInv can also be integrated with existing customized image generation methods, expanding their applications from customized T2I generation to text-driven localized editing.
\item Our experimental results demonstrate that SPDInv effectively mitigates the dependency of inverted noise code on source prompt, significantly reducing the artifacts and inconsistent details in the editing outputs.
\end{itemize}




\section{Related Work}

\subsection{Generative Models for Image Editing}

The rapid advancement of diffusion-based generative models has significantly impacted the field of image and video generation. Large-scale pre-trained models such as Stable Diffusion (SD) \cite{LDM}, GLIDE\cite{nichol2021glide}, Imagen\cite{saharia2022photorealistic}, and DALL$\cdot$E2\cite{ramesh2022hierarchical} have demonstrated powerful image synthesis capability, and have been serving as foundational models for many  downstream tasks. Prompt-to-Prompt (P2P)\cite{Prompt2Prompt} is among the first to utilize SD for complex image editing through text interaction, achieving remarkable localized editing results by controlling attention maps. Subsequent works like pix2pix-zero\cite{zero2023parmar} and plug-and-play (PNP)\cite{plugandplay} offer fine-grained control over textual embedding and spatial features. MasaCtrl \cite{masactrl} manipulates self-attention features for consistent image generation and non-rigid editing simultaneously. By coupling with DDIM inversion, these methods can be used to edit real images, yet they often encounter editing failures due to the instability of DDIM inversion process. 

In addition to localized editing, customized image generation has recently garnered much attention, which aim to generate images with the identity of user-provided object (\eg, dog) unchanged. This can be achieved by fine-tuning parts of a pre-trained diffusion model (\eg, the entire diffusion model\cite{ruiz2023dreambooth,hu2023animate}, cross-attention\cite{hu2021lora,kumari2023multi} module, or text embedding space\cite{gal2022image}) or training an auxiliary module to translate visual content into the text embedding space\cite{wei2023elite,li2023photomaker,wang2024instantid,ye2023ip,mou2023t2i}. While these methods have shown promising results in customized image generation, achieving fine-grained localized editing over customized images remains challenging. Our proposed SPDInv can be integrated into the existing customized generation methods to empower them with localized editing capacity.


\vspace{-2mm}
\subsection{Inversion for Real Image Editing}


Inversion is a crucial step for editing real images in order to achieve reconstruction fidelity and editability. There are mainly four categories of inversions methods. The first category is DDPM \cite{DDPM} based methods, which directly use the DDPM forward equation to obtain the noise code \cite{meng2021sdedit,huberman2023edit,tsaban2023ledits}. However, the obtained noise code by these methods cannot guarantee the reconstruction of the original image. The second category is DDIM \cite{DDIM} optimization-based methods. While DDIM inversion\cite{elarabawy2022direct} can be directly used to obtain the latent noise code, it may fail to reconstruct the image when the CFG parameters are set higher. To enhance reconstruction consistency, methods such as NTI\cite{NTI}, NPI\cite{NPI} and ProxEdit\cite{Prox} have been proposed to optimize the text embedding space. Some other methods have also been proposed to optimize the image latent space \cite{cho2024noise,epstein2024diffusion} or the final noise space\cite{zhang2024real}. The third category is DDIM optimization-free methods, which aim to address the time-consuming issue of optimization-based methods. EDICT\cite{wallace2023edict} and its variants\cite{zhang2023exact} employ auxiliary invertible neural network to compute the inversion path. DirectInv \cite{directinv,xu2023inversion} records the differences between inversion and reconstruction, and then merges the differences into the inference process to ensure high-quality reconstruction. The last category is fine-tuning-based methods, which improve the reconstruction by overfitting the neural network to the given image\cite{shi2023dragdiffusion,li2023stylediffusion,gal2022image} or training auxiliary networks\cite{huang2023reversion,wei2023elite}.

While the above methods can reconstruct well the source image using the inverted noise code, they may generate artifacts and inconsistent details when using the target prompt to edit the image. 
Some works such as AIDI\cite{AIDI} and FPI\cite{meiri2023fixed} have shown the effectiveness of fixed-point constraint in the inversion process. However, they employ the fixed-point iteration to search the solution, which is unstable and sub-optimal. In this work, we reformulate the fixed-point constraint as a loss function and leverage the pre-trained diffusion model to minimize it. Our method significantly reduces the editing artifacts and improves the detail consistency. 

\section{Our Method}

\subsection{Analysis on DDIM Inversion}
\label{sec:preliminary}

\begin{figure}[tb]
  \centering
  \includegraphics[width=\linewidth]{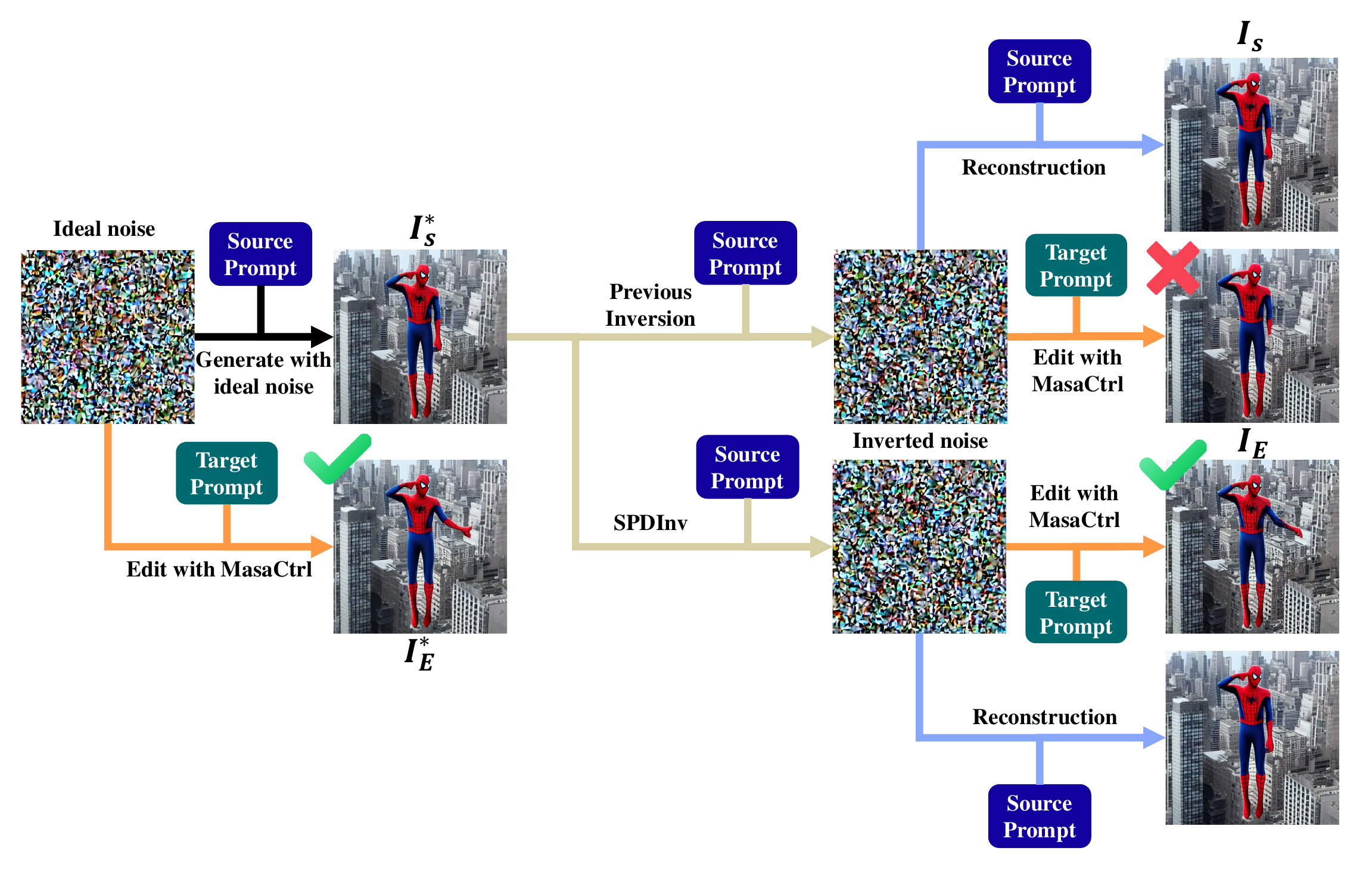}
\vspace{-5mm}
  \caption{ An example of image editing with ideal noise code (left) and inverted noise code (right). Source prompt: A spiderman in the city. Target prompt: A spiderman in the city with his left hand up.}
  \vspace{-3mm}
  \label{fig:failure_case}
\end{figure}

The editing pipeline of most text-driven image editing methods has been depicted in \cref{fig:edit_process}. The given image $z_0$ is first inverted into latent noise $z_T$, which serves as the initial point for reconstruction and editing. DDIM inversion is commonly employed in the existing methods to obtain $z_T$.  
While DDIM inversion and its subsequent methods\cite{NTI,NPI,Prox,directinv} have shown promising results in editing real images, their editing fidelity and flexibility are far from the case if the ideal noise code can be used as the starting point. An example is illustrated in \cref{fig:failure_case}. From a random noise code without any prior knowledge, which can be regarded as an ideal noise, we generate a Spiderman image $I_{S}^*$ with source prompt \textit{"A spiderman in the city"}. With the same ideal noise and by using the MasaCtrl editing engine \cite{masactrl}, we can successfully change the pose of Spiderman with target prompt \textit{"A spiderman in the city with his left hand up"}. However, if we take the inverted noise code of $I_{S}^*$ as the initial point, the edited result, denoted by $I_{E}$, will fail in editing Spiderman's pose (the left hand disappears). This is because during the inversion process, the prior information of source prompt is remained in the inverted noise code. While this prior information facilitates the reconstruction process, it impedes the editing fidelity and flexibility based on the target prompt, resulting in unintended artifacts or content inconsistency.  

Let's make more analyses on the DDIM inversion process. Starting from the pure Gaussian noise $z_T \sim N(0,1)$, DDIM employs a deterministic sampling process\cite{DDIM} $ z_{t-1}
= \frac{\sqrt{\alpha_{t-1}}}{\sqrt{\alpha_t}} z_t + \sqrt{\alpha_{t-1}} \bigg( \sqrt{\frac{1}{\alpha_{t-1}} - 1} - \sqrt{ \frac{1}{\alpha_t} -1}  \bigg) \epsilon_{\theta}(z_t,t,c)$ to generate the image latent code $z_0$. From this sampling equation, we can obtain the ideal inversion equation by using $z_t$ and $z_{t-1}$ with the following equation:
\begin{equation}
    z_t = C_{t,1}*z_{t-1} + C_{t,2}* \redContent{\epsilon_{\theta}(z_t,t,c)},
    \label{eq:ideal_inversion}
\end{equation}
where $C_{t,1}=\frac{ \sqrt{\alpha_t} }{ \sqrt{\alpha_{t-1}} },C_{t,2}=\sqrt{\alpha_t} \big( \sqrt{\frac{1}{\alpha_t} - 1} - \sqrt{\frac{1}{\alpha_{t-1}}-1}  \big)$. The detailed derivation of \cref{eq:ideal_inversion} can be found in the \textbf{supplementary materials}. The inputs of the neural network are supposed to be $z_t$ and $t$. However, based on the assumption that the ODE formula can be reversed within infinitesimally small steps and due to the practical constraint that $z_{t}$ is unavailable during one-step inversion at timestep $t-1$, DDIM inversion uses $(z_{t-1},t-1,c)$, instead of $(z_t,t,c)$,  as the input of the neural network, resulting in the following formula:
\begin{equation}
    z_t = C_{t,1}*z_{t-1} +  C_{t,2}* \epsilon_{\theta}(z_{t-1},t-1,c).
    \label{eq:DDIM_inversion}
\end{equation}

The coupling of DDIM inversion with source prompt is rooted in the use of \cref{eq:DDIM_inversion}. 
In the ideal inversion (\ie, \cref{eq:ideal_inversion}), the input should be $z_t$ and $t$, whereas previous methods have utilized $z_{t-1}$ and $t-1$ as the input (\ie, \cref{eq:DDIM_inversion}). 
However, $z_{t-1}$ is obtained by denoising $z_t$ conditioned with source prompt. Therefore,  \cref{eq:DDIM_inversion} actually introduces source prompt prior into the update. Taking the last inversion step as an example, the neural network should receive a latent $z_T$ a pure noise without any prior information as the input to obtain the ideal inversion code as $z_T= C_{T,1}*z_{T-1} + C_{T,2} *\redContent{\epsilon_{\theta}(z_T,T,c)}$, while most previous methods utilize the latent feature $z_{T-1}$, which is obtained through one-step denoising conditioned on the source prompt, as the input, \ie, $\hat{z}_T= C_{T,1}*z_{T-1} + C_{T,2}* \blueContent{\epsilon_{\theta}(z_{T-1},{T-1},c)}$. The coupling of the inverted noise code and source prompt is not limited to the final inversion step, as the multi-step generation nature of LDM accumulates the divergence between the inverted noise and the ideal noise. This eventually results in the inclusion of source prompt prior in the inverted noise, causing a notable deviation from the ideal noise, \ie, gap of noise $D_{Noi}$ as illustrated in \cref{fig:inversion_process}(a). This deviation ultimately affects the editing stability and fidelity of the given real image.

\subsection{Source Prompt Disentangled Inversion (SPDInv)}

The aforementioned analysis highlights the significance of an ideal noise code, which should be disentangled with the source prompt, in editing a real image with target prompts. In practical applications, however, the corresponding ideal noise for a given image and source prompt is unknown, making it difficult to minimize the divergence between the ideal noise and the inverted noise. Nevertheless, \cref{eq:ideal_inversion} sheds light on the potential solution since it provides a constraint that $z_t$ and $z_{t-1}$ should satisfy at each inversion step. Consequently, narrowing the gap between the inverted and ideal noise codes can be achieved by optimizing $z_t$ and $z_{t-1}$ to meet the constraint of \cref{eq:ideal_inversion}, thereby circumventing the issue of the unavailability of an ideal noise.

The ideal inversion equation in \cref{eq:ideal_inversion} is actually a fixed-point problem, which has been discussed in AIDI\cite{AIDI}. At the beginning of the inversion step, we have $z_{t-1}$ available. Subsequently, by taking $z_t$ as the variable, we can convert \cref{eq:ideal_inversion} into:
\begin{equation}
    x = f_{\theta}(x) \quad where \quad x = z_t,f_{\theta}(x) = C_{t,2} * \epsilon_{\theta}(x,t,c) + C_{t,1}*z_{t-1}.
    \label{eq:fix_point}
\end{equation}
$C_{t,2}$ and $C_{t,1} z_{t-1}$ can be regarded as constants. By optimizing \cref{eq:fix_point}, the obtained $z_t$ will approach to the ideal latent $z_t^*$ in each inversion step. Eventually, the inverted noise code $z_T$ will approach to the ideal noise code $z_T^*$. AIDI \cite{AIDI} performs fixed-round iterations by assigning $f_{\theta}(z_t)$ to $z_t$ to solve a similar fixed-point problem to \cref{eq:fix_point}. However, as LDM utilizes a neural network to map $z_t,t,c$ to noise $\epsilon$, the exact mathematical form of $\epsilon_{\theta}(z_t,t,c)$ cannot be obtained. As a result, the fixed-round iteration may not converge to a good fixed-point. 

\begin{algorithm}
    
	\renewcommand{\algorithmicrequire}{\textbf{Input:}}
	\renewcommand{\algorithmicensure}{\textbf{Output:}}
	\caption{Source Prompt Disentangled Inversion (SPDInv)}
	\label{alg:inversion}
	\begin{algorithmic}[1]
		\REQUIRE Source image latent $z_0$, DDIM steps $T$, source prompt $p_s$, maximal optimization round $K$, threshold $\delta$, learning rate $\eta$.
        \ENSURE Inversion noise $z_T$
        \FOR{$t\leftarrow1$ to T}
            \STATE Get $z_{t}$ from $z_{t-1}$ based on \cref{eq:DDIM_inversion}
            \FOR{$i\leftarrow0$ to K}
                \STATE Calculate $L = ||f_{\theta}(z_{t}) - z_{t}||_2 $ based on \cref{eq:fix_point}
                \STATE Update $z_{t} := z_{t} - \eta \nabla L$
                \STATE \textbf{if} $L<\delta$ \textbf{then} \textit{Break} \textbf{end if}
            \ENDFOR
        \ENDFOR
	\end{algorithmic} 
\end{algorithm}

We thereby propose a Source Prompt Disentangled Inversion (SPDInv) method to mitigate the influence of source prompt on the inverted noise code. We convert each inversion step into a search problem to identify the fixed point. Consequently, we reformulate the search problem into a loss function and leverage a pre-trained diffusion model to facilitate the optimization process. Our approach is straightforward yet effective, requiring only 10 lines of modifications to the existing inversion technique but improving the current state-of-the-arts significantly. Our algorithm is summarized in \textbf{\cref{alg:inversion}}.

Specifically, we utilize \cref{eq:DDIM_inversion} to perform a single-step inversion from $z_{t-1}$ to obtain an initial approximation to $z_t$. At this moment, $z_{t-1}$ and $z_t$ do not satisfy the constraint in \cref{eq:ideal_inversion}. We employ a powerful pre-trained network $\epsilon_{\theta}(z_t,t,c)$ (\ie, the Stable Diffusion 1.4 \cite{LDM}), which is trained on a vast image-text dataset, and leverage its image-text comprehension capability to guide the optimization of \cref{eq:ideal_inversion}. By incorporating $z_t$ into \cref{eq:fix_point}, we transform the searching of $z_t$ into the optimization of the following loss function:
\begin{equation}
    \argminE_{z_t} L = \parallel f_{\theta}(z_t) - z_t \parallel_2.
    \label{eq:loss_term}
    \vspace{-1mm}
\end{equation}
\cref{eq:loss_term} can be minimized by the gradient descent techniques through $z_t := z_t - \eta \nabla L$, where $\eta$ is the learning rate. The pre-trained diffusion model is fixed throughout the optimization process, with only the latent feature $z_t$ being updated. Our experiments demonstrate that our SPDInv method exhibits superior performance compared to AIDI.

Furthermore, we observed that the loss function in \cref{eq:loss_term} converges at varying speeds for different inversion steps $t$. In the early stages of the inversion process, more rounds of optimization are required to meet the fixed-point constraint in \cref{eq:fix_point}. When $t>\frac{T}{2}$, the loss quickly converges within a few rounds. Therefore, in addition to setting a maximal number of rounds $K$ for all inversion steps, we introduce a threshold $\delta$ to control the termination of the optimization process to improve the efficiency of inversion process. 

\subsection{Application to Customized Image Generation}
\label{sec:customized}

\begin{figure}[tb]
  \centering
  \includegraphics[width=0.85\linewidth]{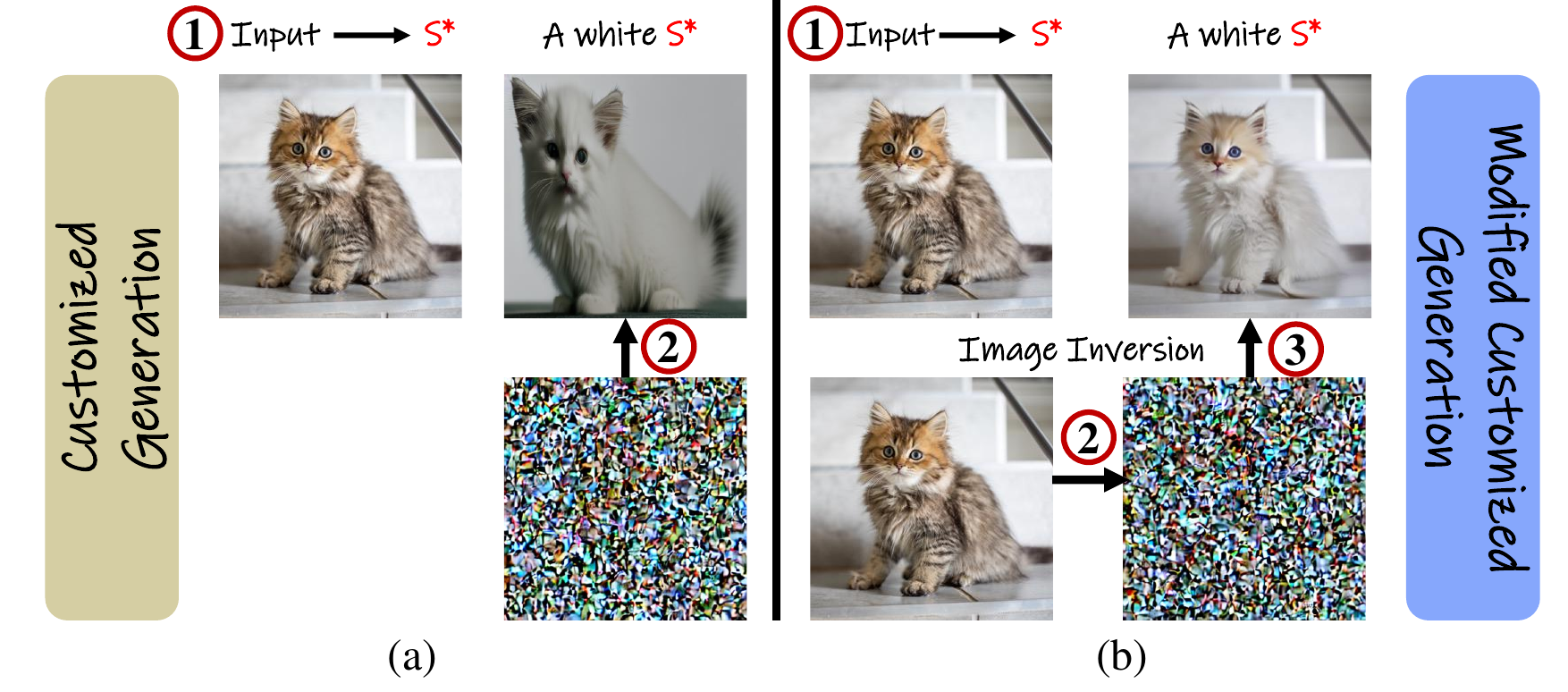}
  \caption{(a) The pipeline of existing customized image generation methods. (b) The new image generation pipeline by integrating our proposed SPDInv method to generate the latent noise code, which can preserve the background of the input image.}
  
  \label{fig:elite_process}
  \vspace{-5mm}
\end{figure}

Customized image generation\cite{tewel2023key,voronov2024loss,choi2023custom} aims to generate new images by using the text concept (denoted by "S*") extracted from the given image together with other input text prompt. 
This can be achieved by fine-tuning parts of a pre-trained diffusion model (\eg, the entire U-Net \cite{ruiz2023dreambooth,hu2023animate}, cross-attention modules\cite{hu2021lora,kumari2023multi}, text embedding space\cite{gal2022image,NTI}) or training an auxiliary module to translate visual contents into the text embedding space \cite{wei2023elite,ye2023ip,mou2023t2i,zhang2023adding}. Especially, recent methods like ELITE\cite{wei2023elite}, PhotoMaker\cite{li2023photomaker}, InstantID\cite{wang2024instantid} can achieve quick customized generation of animals, portraits and other items. However, one of the limitations of customized image generation is that the generated image usually exhibits poor background and layout preservation. One example is depicted in Fig. \ref{fig:elite_process}(a). When we use the state-of-the-art customized image generation method ELITE to change the color of the cat in the given image, a cat with different pose and background can be returned.

To address the above mentioned issue of established customized image generation methods, we can easily  integrate our proposed SPDInv into them to augment their localized editing capabilities. As shown in  \cref{fig:elite_process}, with the existing methods (\eg, ELITE), we can first transform the given image into the text embedding space aligned with text "S*" (\ie, step 1 in Fig. \ref{fig:elite_process}(b)). Then, instead of performing synthesis with a random noise code, we uses SPDInv to invert the image into a noise code (step 2 in Fig. \ref{fig:elite_process}(b)), which works as the key to maintain the layout and background of the input image. Finally, with the inverted noise code and the new text prompt such as "a white S*", we can use pre-trained diffusion models (\eg, stable diffusion v1.4 for ELITE) to generate an image with only the color changed, as depicted in the upper right of Fig. \ref{fig:elite_process}(b). The new pipeline in Fig. \ref{fig:elite_process}(b) extends the capability of existing customized image generation methods to perform high quality localized editing.

\section{Experiment}

\subsection{Experiment Setting}

\textbf{Dataset.} We use two datasets to evaluate our proposed SPDInv method. The first is the \textit{PIE-Bench}\cite{directinv} provided by DirectINV\cite{directinv}, which comprises 700 images with different editing types, including changing object, pose, color, material, background, style, adding and deleting object. In addition, following the setting of NTI\cite{NTI}, we randomly choose 100 images from the \textit{COCO2017}\cite{lin2014microsoft} evaluation dataset without cherry-picking to build another test set. We construct the target prompt for each image with the same editing types as \textit{PIE-Bench}. We call this test set as \textit{TDE-Bench} (\textit{T}ext-\textit{D}riven \textit{E}diting \textit{Bench}mark).  

\textbf{Evaluation Metrics.} Multiple metrics are employed to evaluate the performance of SPDInv from different aspects, including overall structure distance (assessed by DINO score\cite{caron2021emerging}), background preservation (assessed by PSNR, LPIPS, MSE, SSIM), and prompt-image consistency (assessed by CLIP score\cite{radford2021learning}). As in previous methods, DINO score and CLIP score are calculated based on the entire image, while PSNR, LPIPS, MSE, and SSIM are calculated based on the region outside the annotated editing mask.

\textbf{Comparison Methods.} In \cref{sec:exp_inversion}, we compare SPDInv with seven representative and state-of-the-art inversion based editing methods, including DDIM inversion\cite{DDIM}, Null-text inversion (NTI)\cite{NTI}, Negative prompt inversion (NPI)\cite{NPI}, Direct Inversion (DirectINV)\cite{directinv}, ProxEdit\cite{Prox}, Noise Map Guidance (NMG)\cite{cho2024noise}, and AIDI\cite{AIDI}. When the P2P editing engine is used, all these comparison methods can be evaluated. However, many of these methods are inapplicable to the MasaCtrl and PNP editing engine due to the lack of source code. So we can only compare with DirectINV, NMG and AIDI under the MasaCtrl engine,  and compare with DirectINV and AIDI under the PNP engine.

In \cref{sec:exp_customized_generation}, we choose the state-of-the-art customized image generation methods ELITE\cite{wei2023elite} as the baselines to demonstrate the improvement on localized editing brought by our SPDInv. We further compare our improved editing methods with two strong non-inversion based editing methods BlendDM\cite{avrahami2022blended} and InstructP2P\cite{brooks2023instructpix2pix} to demonstrate the effectiveness of SPDInv.

\textbf{Other Settings.} In our experiments, we set the DDIM sampling step as 50, the Classifer Free Guidance (CFG) as 7.5, and other parameters as their default values. The base model utilized is Stable Diffusion v1.4. The experiments and time consumption are tested on RTX3090.

\subsection{Results on Text-Driven Image Editing}
\label{sec:exp_inversion}
\begin{table}[t]
\caption{Performance comparison of different inversion methods under the Prompt-to-Prompt (P2P)\cite{Prompt2Prompt}, Mutual self control (MasaCtrl)\cite{masactrl} and Plug-and-Play (PNP)\cite{plugandplay} editing engines on PIE-Bench. Best and second best metrics are highlighted in \redContent{red} and \blueContent{blue} colors, respectively.}
\vspace{-2mm}
    \label{tab:metrice_edit}
    \centering
    \begin{tabular}{l|c|c|c|c|c|c|c|c}
   \toprule
   \midrule
   \makecell[c]{Inversion} & \makecell[c]{Editing Engine} & \makecell[c]{DINO$\downarrow$ \\ $\times10^3$} & PSNR$\uparrow$ & \makecell[c]{LPIPS$\downarrow$ \\ $\times10^3$} & \makecell[c]{MSE$\downarrow$ \\ $\times10^4$} & \makecell[c]{SSIM$\uparrow$ \\ $\times10^2$} & \makecell[c]{CLIP}$\uparrow$ & \makecell[c]{Inversion \\ times(s)}\\
   \midrule
DDIM& P2P & 69.43 & 17.87 & 208.80 & 219.88 & 71.14 & 25.01 & \redContent{11.55} \\
NTI& P2P & 13.44 & 27.03 & 60.67 & 35.86 & 84.11 & 24.75  & 137.54\\
NPI& P2P & 16.17 & 26.21 & 69.01 & 39.73 & 83.40 & 24.61 & \blueContent{11.75}\\
AIDI& P2P & 12.16 & 27.01 & 56.39 & 36.90 & 84.27 & 24.92 & 87.21\\
NMG & P2P & 23.50 & 25.83 & 81.58 & 107.95 & 82.31 & 24.05 & 16.71 \\
DirectINV & P2P & \blueContent{11.65} & \blueContent{27.22} & 54.55 & 32.86 & 84.76  & \blueContent{25.02} & 19.94\\
ProxEdit & P2P & 11.87 & 27.12 & \blueContent{45.70} & \blueContent{32.16} & \blueContent{84.80} & 24.28 & \blueContent{11.75}\\
\midrule
SPDInv&P2P & \redContent{8.81} & \redContent{28.60} & \redContent{36.01} & \redContent{24.54} & \redContent{86.23} & \redContent{25.26} & 27.04\\
\midrule
\midrule
DDIM & MasaCtrl &28.38 & 22.17 & 106.62 & 86.97 & 79.67 & 23.96 & \redContent{11.55}\\
NMG & MasaCtrl & 40.54 & 20.35 & 127.85 & 135.17 & 77.52 & \blueContent{24.56} & \blueContent{16.71} \\
DirectINV & MasaCtrl & \blueContent{24.70} & \blueContent{22.64} & \blueContent{87.94} & \blueContent{81.09} & \blueContent{81.33} & 24.38 & 19.94\\
AIDI&MasaCtrl & 55.93 & 19.25 & 177.57 & 178.13 & 75.58 & 24.01 & 87.21\\
\midrule
SPDInv&MasaCtrl & \redContent{20.48} & \redContent{24.12} & \redContent{71.74} & \redContent{64.77} & \redContent{82.54} & \redContent{24.61} & 27.04\\
\midrule
\midrule
DDIM&PNP & 28.22 & 22.28 & 113.33 & 83.51 & 79.00 & 24.95  & \redContent{11.55}\\
DirectINV&PNP & \blueContent{24.29} & 22.43 & 106.09 & 80.52 & 79.62 & 25.02 & \blueContent{19.94} \\
AIDI&PNP & 25.36 & \blueContent{23.11} & \blueContent{98.10} & \blueContent{78.19} & \blueContent{80.57} & \blueContent{25.03} & 87.21\\
\midrule
SPDInv&PNP & \redContent{15.58} & \redContent{26.72} & \redContent{91.55} & \redContent{34.69} & \redContent{82.04} & \redContent{25.14} & 27.04\\
   \bottomrule
\end{tabular}

\vspace{-6mm}
\end{table}

\cref{tab:metrice_edit} presents the quantitative results of the competing inversion-based methods on PIE-Bench. We set the maximal optimization round $K=25$ in SPDInv. One can see that SPDInv shows significant improvement over previous methods. When using the P2P editing engine, compared with the second-best methods (DirectINV or ProxEdit), SPDInv achieves visible improvements in DINO score ($24\%\uparrow$), PSNR ($5\%\uparrow$), LPIPS ($21\%\downarrow$), MSE ($13\%\downarrow$), SSIM ($1.43 \uparrow$), and CLIP ($0.24 \uparrow$). The inversion time of SPDInv is longer than DDIM, NPI and ProxEdit, similar to NMG and DirectINV, and much shorter than NTI and AIDI (which also solves a fixed-point problem). With the MasaCtrl and PNP editing engines, SPDInv still shows great improvement over previous methods on most metrics, demonstrating the flexibility and effectiveness of SPDInv.

\begin{figure}[tb]
  \centering
  \includegraphics[width=\linewidth]{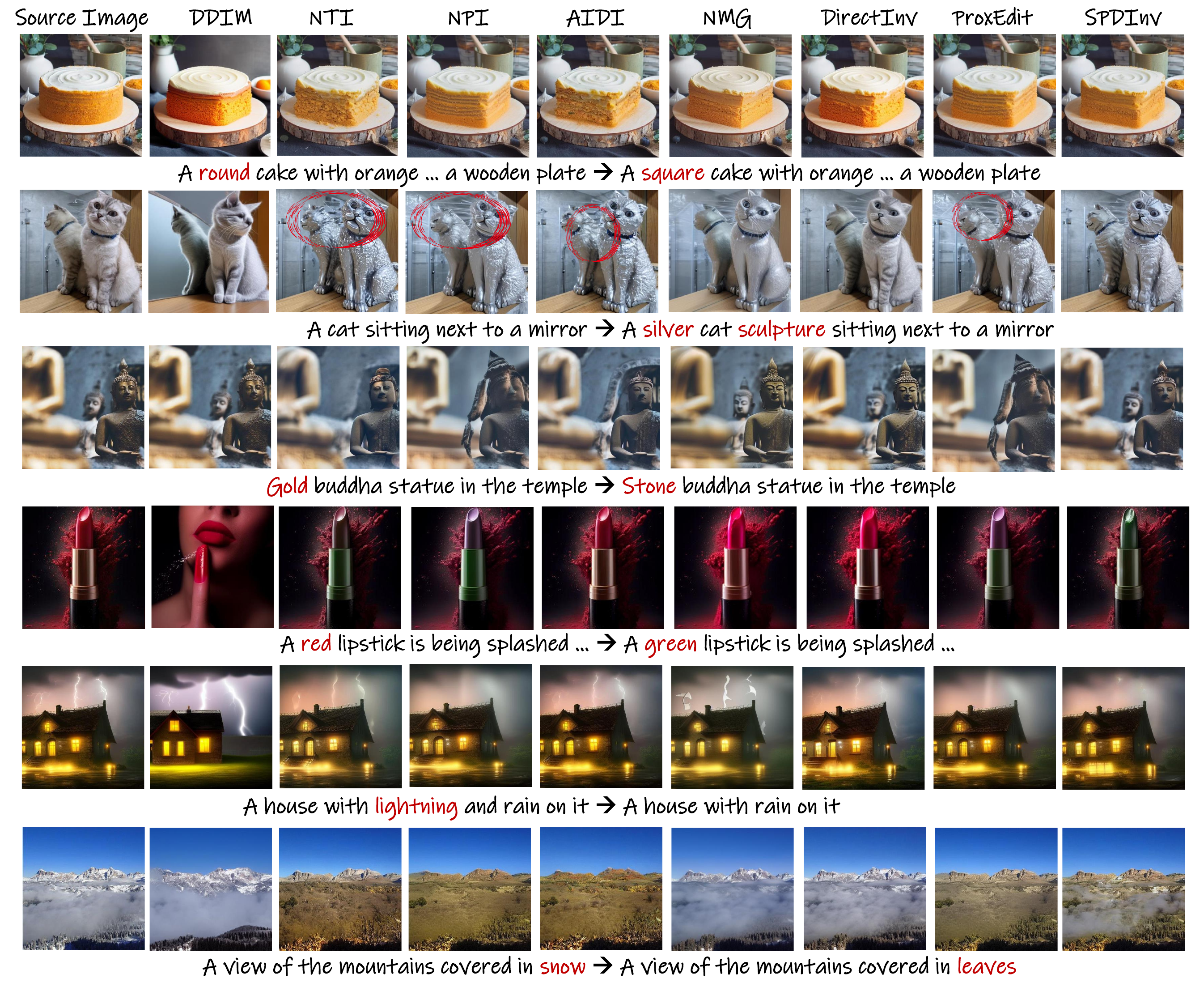}
\vspace{-6mm}
  \caption{Visual comparison of different editing methods with P2P on PIEBench.}
  \label{fig:visual_result}
  \vspace{-7mm}
\end{figure}

The visual comparison results are illustrated in \cref{fig:visual_result} using the P2P engine. We can see that DDIM inversion always exhibits poor content consistency to the input target prompts. While the other competing methods show good editing performance on image cake (row 1), NTI, NPI, AIDI and ProxEdit suffer from artifacts in editing the image cat (row 2), detail inconsistency in editing the images statue (row 3), and fail in editing lipstick (row 5) and lightning (row 6). NMG and DirectInv fail to follow the editing instruction on image lipstick (row 4), lightning (row 5) and mountain (row 6). On the contrary, our SPDInv achieves successful editing in all these cases. More visual comparisons using MasaCtrl and PNP editing engines can be found in the \textbf{supplementary material}.

Due to the limited space, we put the experimental results on TDE-Bench in the \textbf{supplementary material}. Similar conclusions can be made. 

\subsection{Results on Localized Editing with Customized Generation}
\label{sec:exp_customized_generation}
\begin{figure}[tb]
  \centering
  \includegraphics[width=\linewidth]{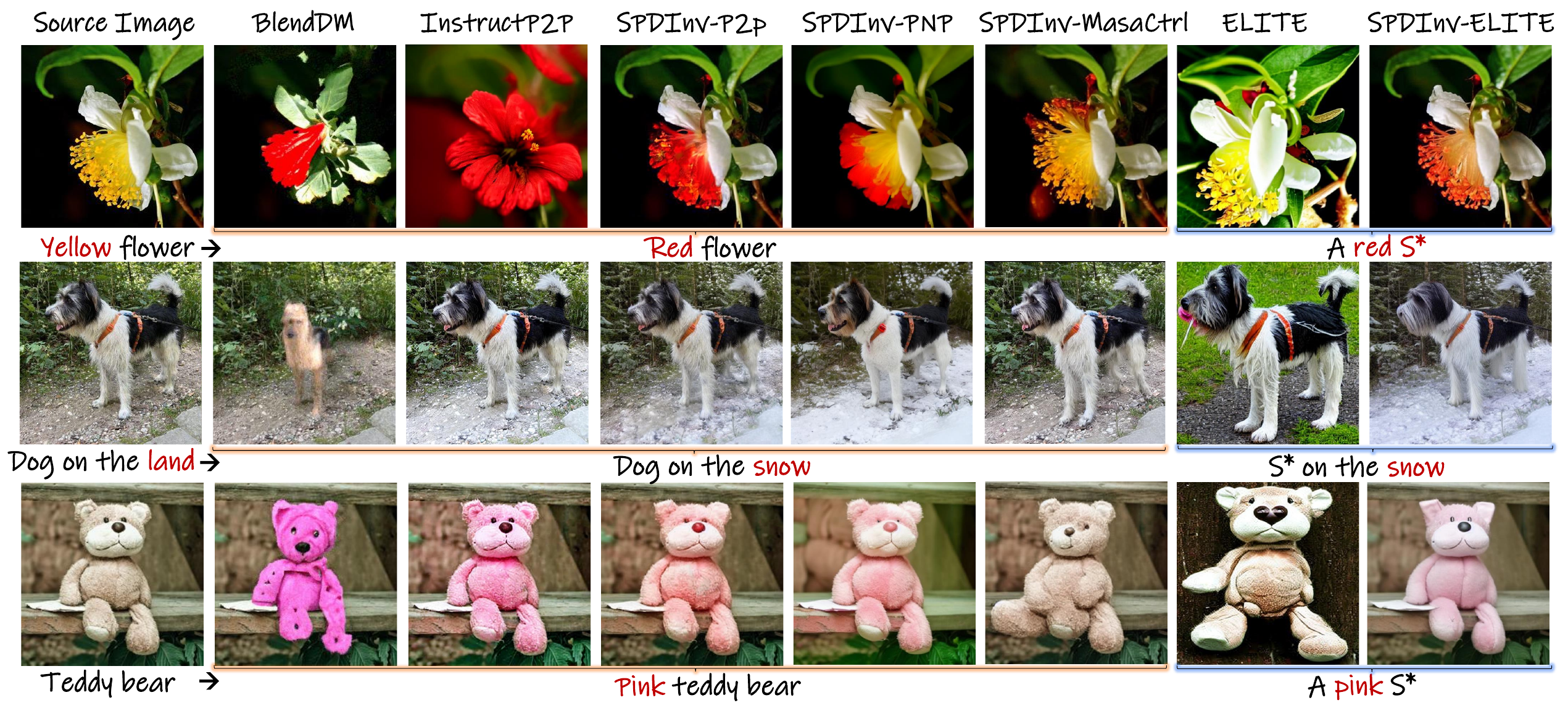}
  \vspace{-5mm}
  \caption{Visual comparisons of localized image editing by different methods.}
  \label{fig:elite_visual}
  \vspace{-1mm}
\end{figure}

\begin{table}[t]
\caption{Evaluation on localized and customized image editing. }
\vspace{-2mm}
    \label{tab:ELITE}
    \centering
    \begin{tabular}{l|c|c|c|c|c|c}
   \toprule
   \midrule
   \makecell[c]{Edit Engine} & $\text{DINO}_{\times10^3}\downarrow$ & PSNR$\uparrow$ & $\text{LPIPS}_{\times10^3}\downarrow$ & $\text{MSE}_{\times10^4}\downarrow$ & $\text{SSIM}_{\times10^2}\uparrow$ & CLIP$\uparrow$ \\
    \midrule
    BlendDM & 59.21 & 15.51 & 244.07 & 306.75 & 67.45 & 20.21 \\
    InstructP2P & 155.49 & 18.19 & 161.20 & 362.01 & 78.05 & 20.06 \\
    ELITE & 148.37 & 14.83 & 201.94 & 359.58 & 67.62 & 15.72 \\
    \midrule
    SPDInv-P2P & 14.77 & 26.59 & 56.30 & 42.58 & 91.20 & 18.63 \\
    SPDInv-MasaCtrl & 31.78 & 22.04 & 91.59 & 71.79 & 87.34 & 16.04 \\
    SPDInv-PNP & 31.33 & 23.23 & 83.63 & 69.79 & 86.90 & 18.83 \\
    \midrule
    SPDInv-ELITE & 21.23 & 24.14 & 74.36 & 48.73 & 88.90 & 19.18 \\
   \bottomrule
\end{tabular}
\vspace{-5mm}
\end{table}

As described in \cref{sec:customized}, SPDInv can be integrated into the existing customized image editing methods such as ELITE \cite{wei2023elite} to endow them the localized editing capability. In this section, we integrate SPDInv into ELITE and name the resulted method as SPDInv-ELITE. We then verify the performance of SPDInv-ELITE on localized and customized image editing by using the test dataset provided by ELITE, which includes 20 subjects and the corresponding masks.

\cref{tab:ELITE} presents the quantitative results. In addition to ELITE, we also provide the results of another two popular non-inversion editing methods BlendDM \cite{avrahami2022blended} and InstructP2P \cite{brooks2023instructpix2pix}. The results of P2P, PNP and MasaCtrl coupled with SPDInv are also listed. 
We see that the original ELITE performs worse than BlendDM and InstructP2P because it has poor background preservation capability. However, upon integration with SPDInv, significant improvements are achieved in all metrics, including DINO score ($85\%\uparrow$), PSNR ($62\%\uparrow$), LPIPS ($63\%\downarrow$), MSE ($86\%\downarrow$), SSIM ($21.28 \uparrow$), and CLIP (3.46 $\uparrow$). Furthermore, SPDInv-ELITE shows superior performance to SPDInv-PNP and SPDInv-MasaCtrl and comparable performance to SPDInv-P2P across the evaluation metrics. 

The visual comparisons are depicted in \cref{fig:elite_visual}. We see that the original ELITE can preserve the identity of object in the image, but suffers from the preservation of background and layout. Non-inversion editing method BlendDM achieves good background preservation but compromises the identities of flower (row 1) and teddy bear (row 3). InstructP2P shows good editing results in some cases (rows 2 and 3) but tends to change the identity and background in other cases (row 1). In contrast, SPDInv-ELITE enables localized and customized editing with improved identity and background preservation. Additionally, SPDInv-ELITE outperforms MasaCtrl in identity and background preservation, and exhibits competitive visual results with P2P and PNP but superior performance in background preservation (row 2).

\subsection{Ablation Study on Hyper-parameter Selection}

There are four hyper-parameters in our SPDInv algorithm, \ie,  $\delta,\eta,K,T$. Via extensive experiments, we empirically set them to  $\delta=5e^{-6},\eta=0.001,K=25,T=50$. Based on these default settings, in this section we conduct ablation experiments to investigate their effects on the final results by altering only one parameter while keeping the others fixed. Specifically, we select $K \in \{5,25,50\}, \delta \in \{5e^{-4},5e^{-5},5e^{-6},5e^{-7\}}, \eta \in \{0.005,0.001,0.01,0.1\}$ and $T\in \{10,50,75,100\}$. The first subset of PIE-Bench is employed in the ablation study.

The results are shown in \cref{tab:Ablation}. First, it is not a surprise that a higher value of $K$ leads to improvements in PSNR, LPIPS, MSE, SSIM and DINO metrics because more iterations are used to solve the fixed-point problem. Nonetheless, there is a slight decline in the CLIP metric, and it requires more computational cost. Therefore, we choose $K=25$ as the default value. Second, for the threshold $\delta$, SPDInv achieves the best performance at $\delta=5e^{-6}$. Further decrease of $\delta$ yields marginal improvements but results in increased inversion time. Third, for the learning rate $\eta$, SPDInv performs the best when $\eta=0.001$. Though slight enhancements in CLIP metrics can be achieved with $\eta=0.005$ or $\eta=0.01$, this comes at the price of deterioration of other metrics. Thus $\eta=0.001$ is selected as our default setting. Finally, for the DDIM sampling steps $T$, we select $T=50$ as the default value due to its balanced performance across all metrics. In addition, $T=50$ is also the default setting in all the baselines mentioned in \cref{sec:exp_inversion}. 

\begin{table}[tp]\scriptsize
\caption{Ablation study on the hyper-parameters of SPDInv with PIE-Bench.}
    \label{tab:Ablation}
    \centering
    
    \begin{tabular}{l|c|c|c|c|c|c}
   \toprule
   \midrule
    \makecell[c]{Hyper parameter} & $\text{DINO}_{\times10^3}\downarrow$ & PSNR$\uparrow$ & $\text{LPIPS}_{\times10^3}\downarrow$ & $\text{MSE}_{\times10^4}\downarrow$ & $\text{SSIM}_{\times10^2}\uparrow$ & CLIP$\uparrow$ \\
    \midrule
    $K=5$ & 8.52 & 31.49 & 22.31 & 10.42 & 90.21 & 26.70 \\
    $K=25$ & 8.43 & 31.61 & 21.70 & 10.12 & 90.28 & 26.67  \\
    $K=50$ & 7.41 & 32.12 & 20.55 & 9.23 & 90.56 & 26.32  \\
    \midrule
    $\delta=5e^{-5}$ & 9.00 & 29.61 & 29.24 & 16.02 & 89.83 & 26.85  \\
    $\delta=5e^{-6}$ & 8.43 & 31.61 & 21.70 & 10.12 & 90.28 & 26.67  \\
    $\delta=5e^{-7}$ & 8.59 & 31.65 & 22.30 & 10.23 & 90.22 & 26.67  \\
    \midrule
    $\eta=0.005$ & 10.39 & 31.08 & 23.60 & 11.33 & 90.00 & 26.87  \\
    $\eta=0.001$ & 8.43 & 31.55 & 22.13 & 10.28 & 90.10 & 26.67  \\
    $\eta=0.01$ & 12.65 & 30.52 & 30.20 & 14.63 & 89.42 & 26.72  \\
    $\eta=0.1$ & 48.91 & 22.82 & 130.98 & 149.87 & 80.89 & 23.26  \\
    \midrule
    $T=10$ & 7.20 & 31.78 & 20.65 & 10.02 & 90.40 & 25.88 \\
    $T=50$ & 8.43 & 31.61 & 21.70 & 10.12 & 90.28 & 26.67  \\
    $T=75$ & 11.26 & 31.27 & 22.55 & 10.74 & 90.11 & 27.05 \\
    \midrule
    \textbf{Default} & \textbf{8.43} & \textbf{31.61} & \textbf{21.70} & \textbf{10.12} & \textbf{90.28} & \textbf{26.67}  \\
   \bottomrule
\end{tabular}
\vspace{-6mm}
\end{table}

\section{Conclusion}
We proposed SPDInv, a novel inversion method designed to enhance the editability of text-driven image editing. To make the inverted noise code as independent as possible to the source prompt so that the image can be better edited with the target prompt, we incorporated a fixed-point constraint to the the inversion process, and showed that this fixed-point constraint could be effectively transformed into a loss function. Consequently, we utilized a pre-trained diffusion model to minimize this loss, and successfully disentangled the inverted noise code with source prompt, significantly improving the editing fidelity and flexibility. Additionally, by integrating SPDInv into customized image generation methods, we enhanced their localized editing capabilities. Experimental results demonstrated the superior performance of SPDInv on benchmark datasets. 

SPDInv has some limitations. First, it relies on the existing editing engines (\ie, P2P, PND, MasaCtrl) to edit images and thus inherits the limitations of them, such as low successful rate in adding and drop contents. 
Second, while SPDInv yields promising results in editing animals, foods and routine items, it still faces difficulties in editing portraits. 
Finally, SPDInv indeed narrows the gap between the inverted noise code and the ideal noise code, but it does not completely eliminate it.
In the future, we will on one hand further improve the inversion process of SPDInv, and on the other hand design new editing pipelines to improve the stability and robustness of image editing results.  

\clearpage

\appendix

\title{Supplementary Material}
\author{ }
\institute{ }

\maketitle

In this supplementary file, we provide the following materials:

\begin{itemize}
    \item Noise gap $D_{noi}$ reduction by our SPDInv (referring to Sec. 1 and Fig. 2 in the main paper);
    \item Derivation of Eq. (1) (referring to Eq. (1) in Sec. 3.1 of the main paper);
    \item Experimental results on TDE-Bench (referring to Sec. 4.2 in the main paper;
    \item More visual comparisons between different methods using P2P, MasaCtrl and PNP editing engines (referring to Sec. 4.2 in the main paper);
    \item More localised editing results with customized generation methods (referring to Sec. 4.3 in the main paper);
    \item Failure cases (referring to Sec. 5 in the main paper).
\end{itemize}

\section{Reduction of Noise Gap by SPDInv}

We conduct experiments to evaluate the noise gap reduction by the proposed SPDInv (please refer to Fig. 2 in the main paper and the analysis in the introduction section for details of noise gap $D_{noi}$). We first utilize the captions extracted from the COCO2017 evaluation dataset\cite{lin2014microsoft} to generate 100 images by Stable Diffusion V1.4. We then record the initial noises $z_T$ and all the latent features $z_t, t\in[0,T)$, which can be regarded as the ground-truth features of the generated images (please refer to the generation path in Fig. 2 of our main paper). Since previous text-driven image editing methods mostly employ DDIM inversion to compute the inverted noise code, we thus compare DDIM inversion with our SPDInv on their inversion performance using the generated ground-truth features and conduct qualitative and quantitive experiments.

\begin{figure}[h]
  \centering
  \includegraphics[width=0.90\linewidth]{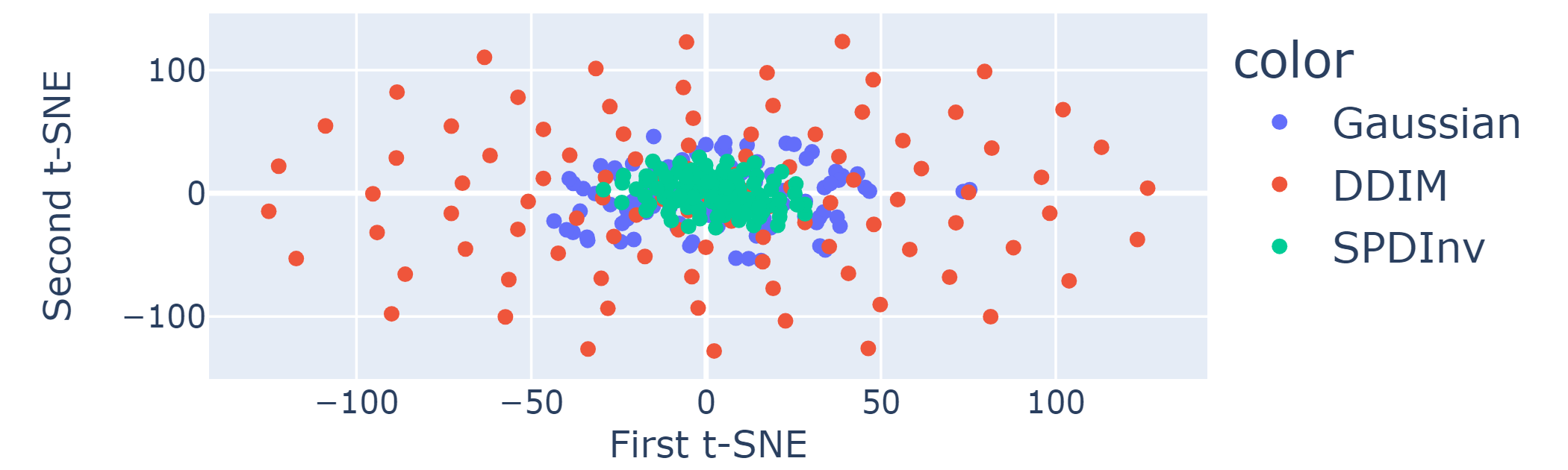}
   \caption{Visualization of inverted noise codes.}
   \label{fig:visual}
\end{figure}

For qualitative evaluation, we utilize t-SNE to reduce the dimension of inverted noise codes and visualize them in \cref{fig:visual}. One can clearly see that the noise codes acquired through SPDInv exhibit a distribution more akin to Gaussian noise than DDIM inversion.

\begin{figure}[t]
  \centering
  \includegraphics[width=\linewidth]{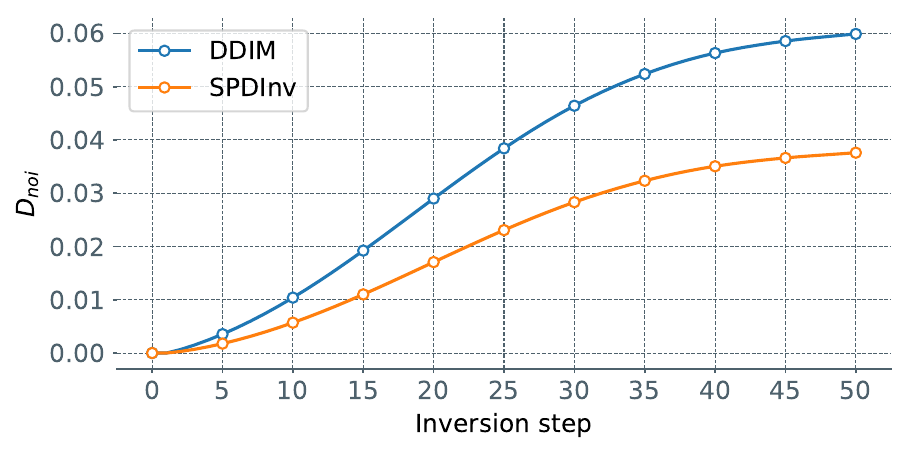}
\vspace{-7mm}
  \caption{The noise gap $D_{noi}$ of inverted noise codes by DDIM inversion and our SPDInv using 100 generated images with captions extracted from COCO2017.}
  \label{fig:noise_gap}
\end{figure}

For quantitive evaluation, we first record all inverted latent features during the inversion process. Then for each inversion step $t$, we can calculate the mean square error (MSE) between the ground-truth and inverted latent features. Finally, the MSEs of all the 100 images are averaged as the $D_{noi}$, which is illustrated in \cref{fig:noise_gap}. Compared with DDIM inversion, our SPDInv reduces the noise gap from 0.06 to 0.04 ($36\%\downarrow$) after 50 inversion steps. Based on the same setting, we further calculate the clip score between the Gaussian noise, DDIM inverted noise, SPDInv inverted noise and the source prompt. The results are shown in \cref{tab:clip_noise}. We see that SPDInv can reduce the clip score by 10\% compared with DDIM inversion, and it is closer to the clip score calculated by pure Gaussian noise. Owe to the reduced noise gap, our SPDInv can effectively reduce the effect of source prompt on the inverted noise code, and consequently reduce the editing artifacts.

\begin{table}[t]
\caption{Clip score between noise codes and source prompt.}
    \label{tab:clip_noise}
    \centering
    \begin{tabular}{c|c|c|c}
   \toprule
   \midrule
   Metrics & Gaussian Noise & DDIM & SPDInv \\
    \midrule
   Clip Score & 12.66 & 14.55 & 13.18 \\
   \bottomrule
\end{tabular}
\end{table}

\section{Derivation of Eq. (1) in the Main Paper}
\label{sec:proof_of_prior}

According to DDIM \cite{DDIM}, the deterministic sampling equation (please refer to the proof of DDIM\cite{DDIM}) is:
\begin{equation}
    z_{t-1}
= \frac{\sqrt{\alpha_{t-1}}}{\sqrt{\alpha_t}} z_t + \sqrt{\alpha_{t-1}} \bigg( \sqrt{\frac{1}{\alpha_{t-1}} - 1} - \sqrt{ \frac{1}{\alpha_t} -1}  \bigg) \epsilon_{\theta}(z_t,t,c),
\label{eq:DDIM}
\end{equation}
where $z_t$ is used as the input of the neural network $\epsilon_{\theta}$. $\alpha_t,\alpha_{t-1},c$ can be regarded as constants. $z_{t-1}$ is the output less noisy code by \cref{eq:DDIM}. If we want to put $z_{t}$ on the left side of the equation as the output, we have the following step by step derivations:
\begin{equation}
\begin{split}
    z_{t-1}
&= \frac{\sqrt{\alpha_{t-1}}}{\sqrt{\alpha_t}} z_t + \sqrt{\alpha_{t-1}} \bigg( \sqrt{\frac{1}{\alpha_{t-1}} - 1} - \sqrt{ \frac{1}{\alpha_t} -1}  \bigg) \epsilon_{\theta}(z_t,t,c),  \\
\frac{\sqrt{\alpha_{t-1}}}{\sqrt{\alpha_t}} z_t 
&= z_{t-1} - \sqrt{\alpha_{t-1}} \bigg( \sqrt{\frac{1}{\alpha_{t-1}} - 1} - \sqrt{ \frac{1}{\alpha_{t}} -1}  \bigg) \epsilon_{\theta}(z_t,t,c),  \\
    \frac{\sqrt{\alpha_{t-1}}}{\sqrt{\alpha_t}} z_t 
&= z_{t-1} + \sqrt{\alpha_{t-1}} \bigg( \sqrt{\frac{1}{\alpha_{t}} - 1} - \sqrt{ \frac{1}{\alpha_{t-1}} -1}  \bigg) \epsilon_{\theta}(z_t,t,c),  \\
    z_t &= \frac{ \sqrt{\alpha_t} }{ \sqrt{\alpha_{t-1}} } z_{t-1} + \sqrt{\alpha_t} \bigg( \sqrt{\frac{1}{\alpha_t} - 1} - \sqrt{\frac{1}{\alpha_{t-1}}-1}  \bigg) \redContent{\epsilon_{\theta}(z_t,t,c)},
    \label{eq:SP_ideal_inversion}
\end{split}
\end{equation}
where $z_{t-1}$ is used to calculate $z_t$ to achieve the goal of image inversion. In \cref{eq:SP_ideal_inversion}, \redContent{$(z_t,t,c)$} are supposed to be the input of neural network \redContent{$\epsilon_{\theta}$}. However, previous methods mostly utilized \blueContent{$(z_{t-1},t-1,c)$} as the input of \redContent{$\epsilon_{\theta}$}, coupling the inverted noise code with source prompt (please refer to Sec. 3.1 in main paper for our detail analysis). 

\section{Results on TDE-Bench Dataset}
\begin{figure}[t]
  \centering
  \includegraphics[width=\linewidth]{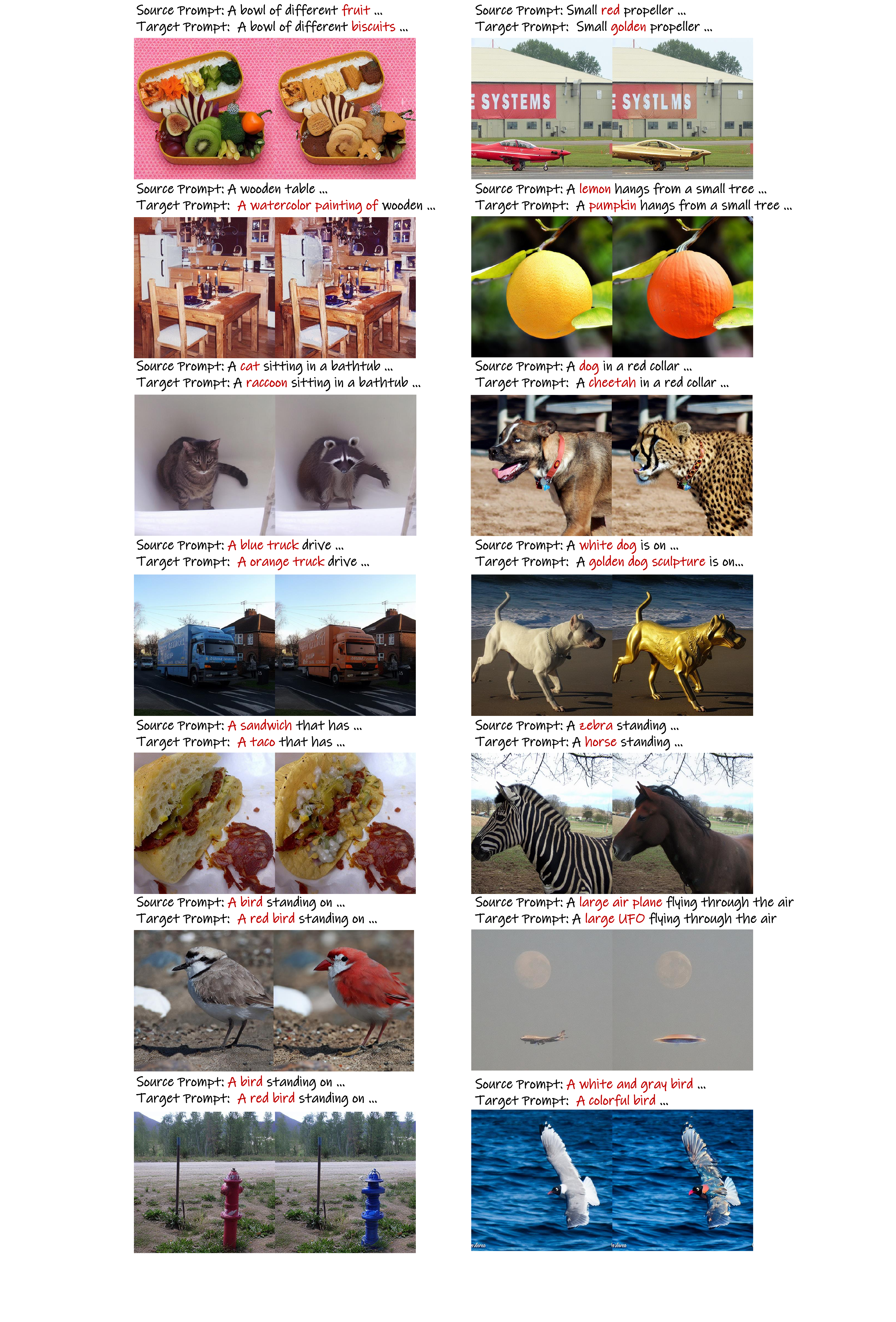}
\vspace{-2mm}
  \caption{Results of SPDInv on TDE-Bench.}
  \label{fig:TDE_cases}
\end{figure}

\cref{tab:TDEbench_result} presents the quantitative results of the competing inversion-based methods on the TDE-Bench. The parameter settings are the same as the Tab. 1 in the main paper. When using the P2P editing engine, compared with the second-best methods (DirectINV or DDIM), SPDInv achieves visible improvements in DINO score ($12\%\uparrow$), PSNR ($1.7\%\uparrow$), LPIPS ($10\%\downarrow$), MSE ($7.7\%\downarrow$), SSIM ($0.41 \uparrow$), and CLIP ($0.03 \uparrow$). When the MasaCtrl and PNP editing engines are used, greater improvement on most metrics can be achieved by using SPDInv.
\cref{fig:TDE_cases} visualizes some editing results, including changing texture (fruit, lemon, sandwich), content (cat, dog, zebra, airplane), color (bird, truck, airplane), and style (room) on TDE-Bench.

\begin{table}[t]
\caption{Performance comparison of different inversion methods under the Prompt-to-Prompt (P2P)\cite{Prompt2Prompt}, Mutual self control (MasaCtrl)\cite{masactrl} and Plug-and-Play (PNP)\cite{plugandplay} editing engines on TDE-Bench. Best and second best metrics are highlighted in \redContent{red} and \blueContent{blue} colors, respectively.}
\vspace{-2mm}
    \label{tab:TDEbench_result}
    \centering
    \begin{tabular}{l|c|c|c|c|c|c|c|c}
   \toprule
   \midrule
   \makecell[c]{Inversion} & \makecell[c]{Editing Engine} & \makecell[c]{DINO$\downarrow$ \\ $\times10^3$} & PSNR$\uparrow$ & \makecell[c]{LPIPS$\downarrow$ \\ $\times10^3$} & \makecell[c]{MSE$\downarrow$ \\ $\times10^4$} & \makecell[c]{SSIM$\uparrow$ \\ $\times10^2$} & \makecell[c]{CLIP}$\uparrow$ & \makecell[c]{Inversion \\ times(s)}\\
   \midrule
DDIM& P2P & 77.50 & 23.23 & 101.23 & 115.33 & 84.83 & \blueContent{25.73} & \redContent{11.55} \\
NTI& P2P & 17.32 & 28.54 & 58.14 & 44.11 & 88.69 & 25.70  & 137.54\\
NPI& P2P & 20.74 & 28.10 & 63.67 & 48.91 & 88.42 & 25.59 & \blueContent{11.75}\\
AIDI& P2P & 15.20 & 28.85 & 59.14 & 46.54 & 88.83 & 25.57 & 87.21\\
NMG & P2P & 19.18 & 29.08 & 52.78 & 37.99 & 89.29 & 25.23 & 16.71 \\
DirectINV & P2P & \blueContent{12.75} & \blueContent{29.17} & \blueContent{49.41} & \blueContent{37.39} & \blueContent{89.48} & 25.57 & 19.94\\
ProxEdit & P2P & 18.67 & 28.60 & 56.54 & 42.89 & 88.91 & 25.60 & \blueContent{11.75}\\
\midrule
SPDInv&P2P & \redContent{11.23} & \redContent{29.69} & \redContent{44.25} & \redContent{34.50} & \redContent{89.89} & \redContent{25.76} & 27.04\\
\midrule
\midrule
DDIM & MasaCtrl & 82.68 & 22.16 & 105.83 & 137.55 & 84.33 & \redContent{26.75} & \redContent{11.55}\\
NMG & MasaCtrl & \blueContent{42.84} & \blueContent{24.75} & \blueContent{70.28} & \blueContent{81.67} & \blueContent{87.37} & 25.45 & \blueContent{16.71} \\
DirectINV & MasaCtrl & 59.29 & 24.46 & 70.47 & 82.74 & 87.33 & 26.13 & 19.94\\
AIDI & MasaCtrl & 80.55 & 21.97 & 106.30 & 140.01 & 84.35 & 26.30 & 87.21\\
\midrule
SPDInv & MasaCtrl & \redContent{22.93} & \redContent{27.96} & \redContent{45.88} & \redContent{42.33} & \redContent{89.32} & \blueContent{26.32} & 27.04\\
\midrule
\midrule
DDIM& PNP & 30.66 & 26.14 & 71.41 & 64.03 & 87.91 & \redContent{26.79}  & \redContent{11.55}\\
DirectINV& PNP & \blueContent{27.79} & 26.17 & \blueContent{69.07} & 63.13 & 88.01 & 26.61 & \blueContent{19.94} \\
AIDI& PNP & 28.16 & \blueContent{26.75} & 69.57 & \blueContent{59.56} & \blueContent{88.18} & 26.68 & 87.21\\
\midrule
SPDInv&PNP & \redContent{20.03} & \redContent{29.73} & \redContent{55.94} & \redContent{25.55} & \redContent{89.07} & \blueContent{26.72} & 27.04\\
   \bottomrule
\end{tabular}

\vspace{-6mm}
\end{table}

\section{More Visual Comparisons between Different Methods using P2P, MasaCtrl and PNP Editing Engines}
\subsubsection{Results with P2P:} 
\begin{figure}[tb]
  \centering
  \includegraphics[width=\linewidth]{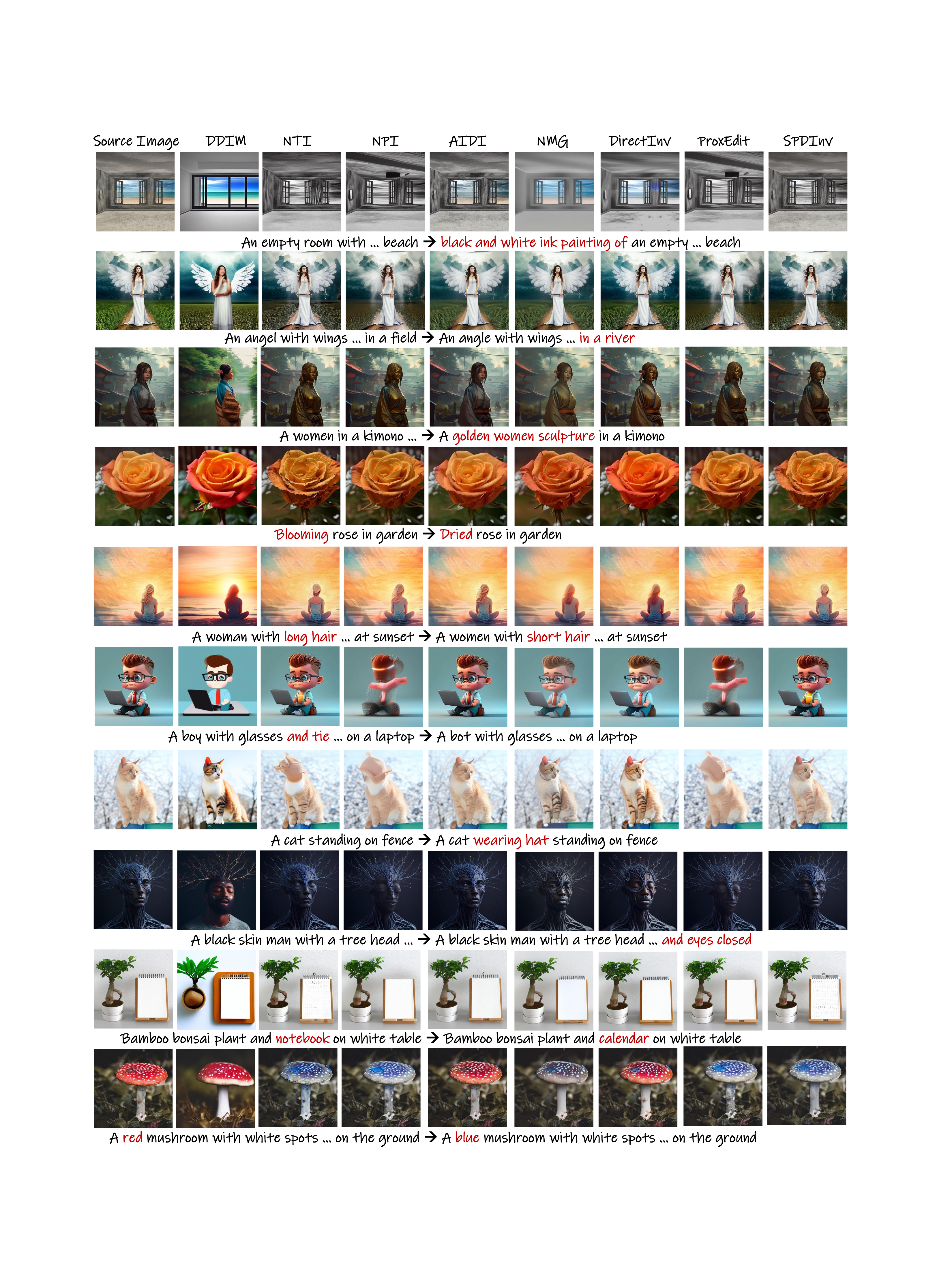}
\vspace{-6mm}
  \caption{More Visual comparisons of different editing methods with P2P.}
  \label{fig:SP_visual_result}
\end{figure}

In \cref{fig:SP_visual_result}, we provide more visual comparisons of competing text-driven image editing methods with the P2P engine. Similar to the results in the main paper, DDIM struggles to preserve the details in most cases. NTI encounters the problem of content inconsistency (women in row 3) and collapse (boy in row 6). NPI and ProxEdit collapse in editing boy (row 6) and cat (row 7). They may also change details in some cases (room layout in row 1, wings in row 2, face in row 8). AIDI, NMG and DirectINV fail in editing rose (row 4), tie (row 5), mushroom (row 10), while NMG and DirectINV change the human face in row 7. On the contrary, SPDInv achieves successful editing in most cases.

\subsubsection{Results with MasaCtrl:} 
\begin{figure}[tb]
  \centering
  \includegraphics[width=\linewidth]{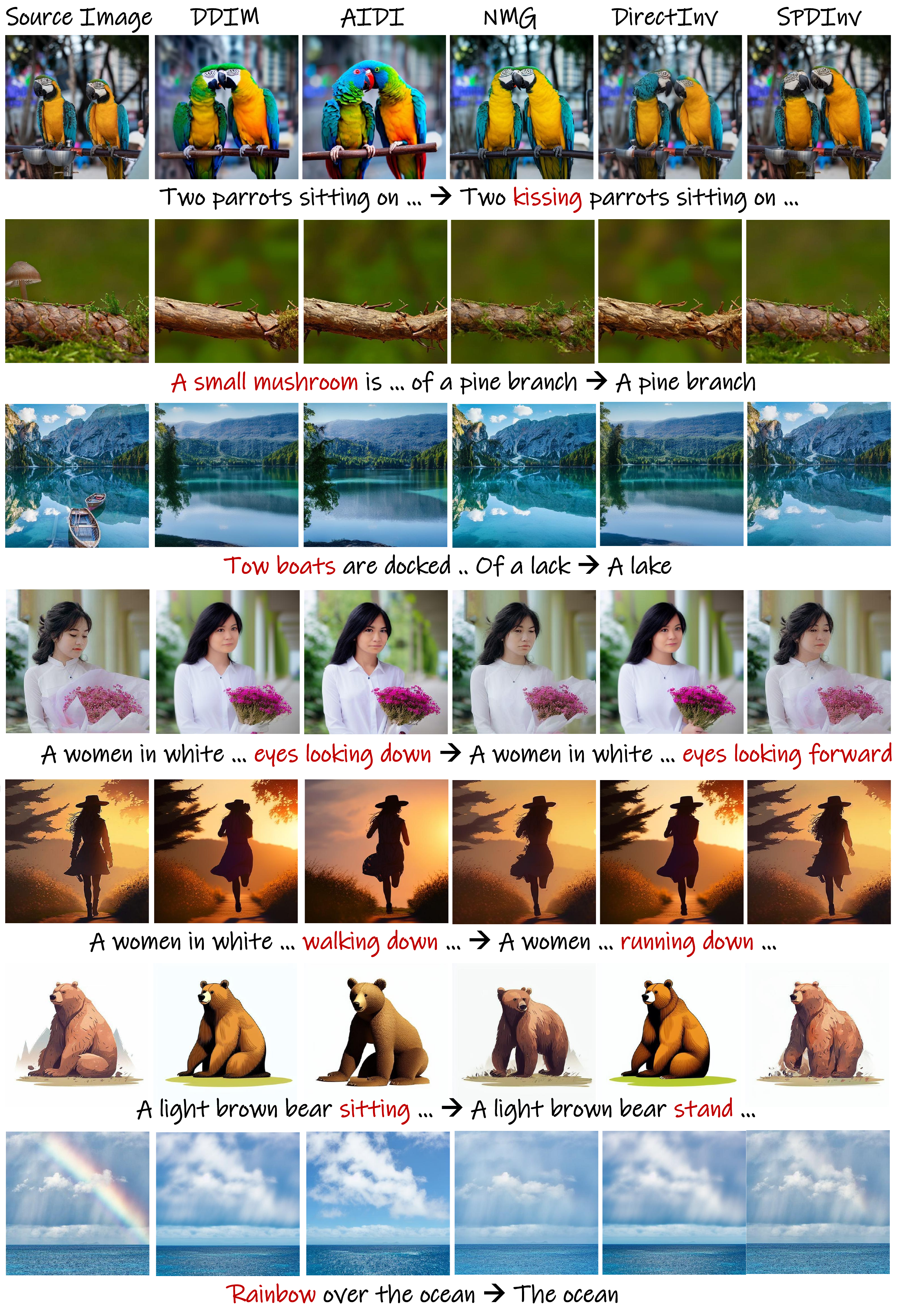}
\vspace{-6mm}
  \caption{Visual comparisons of different editing methods with MasaCtrl.}
  \label{fig:SP_masa}
\end{figure}

\cref{fig:SP_masa} provides the visual results with MasaCtrl editing engine. Different from P2P, MasaCtrl prefers editing nonrigid objects, such as changing pose or removing items, because it can drastically change the structure of the main subject. All competing methods show good editing performance on removing mushroom (row 2), boat (row 3). However, DDIM, AIDI and DirectINV lose details of mountain (row 3) or bear (row 6). NMG shows competitive performance in most cases. Nevertheless, SPDInv exhibits better detail preservation on branch in row 1, and reflection in row 3.

\subsubsection{Results with PNP:} 
\begin{figure}[tb]
  \centering
  \includegraphics[width=\linewidth]{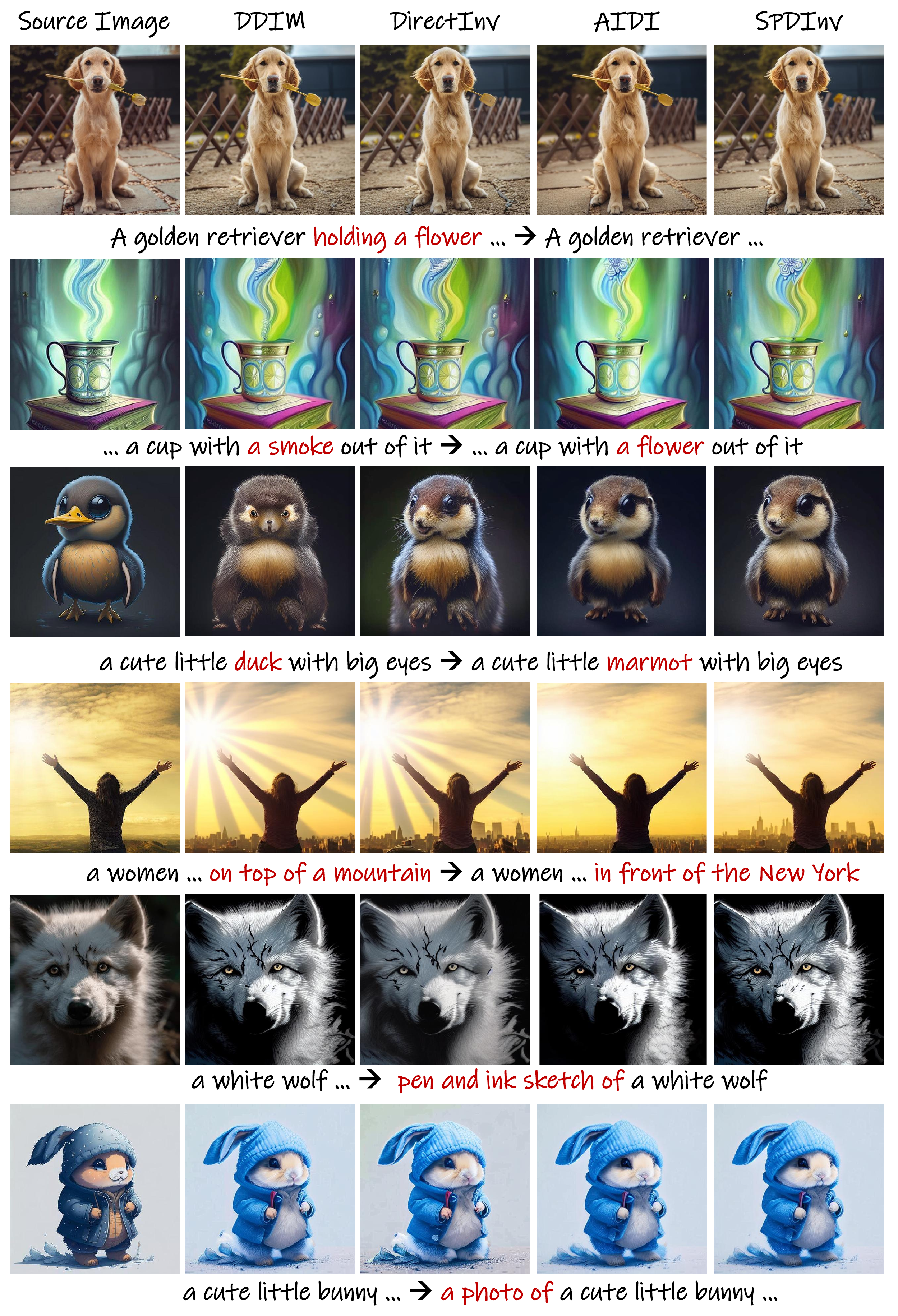}
\vspace{-6mm}
  \caption{Visual comparisons of different editing methods with PNP.}
  \label{fig:SP_PNP}
\end{figure}

\cref{fig:SP_PNP} provides the results with PNP engine. DDIM and DirectINV do not perform very well on images such as dog (row 1) and mountain (row 4), and show inconsistency during editing the duck (row 3) and bunny (row 6). AIDI fails in removing the flower in row 1. SPDInv achieves successful editing in all these cases.

\section{Localized Editing with Customized Generation Methods}
\begin{figure}[t]
  \centering
  \includegraphics[width=0.95\linewidth]{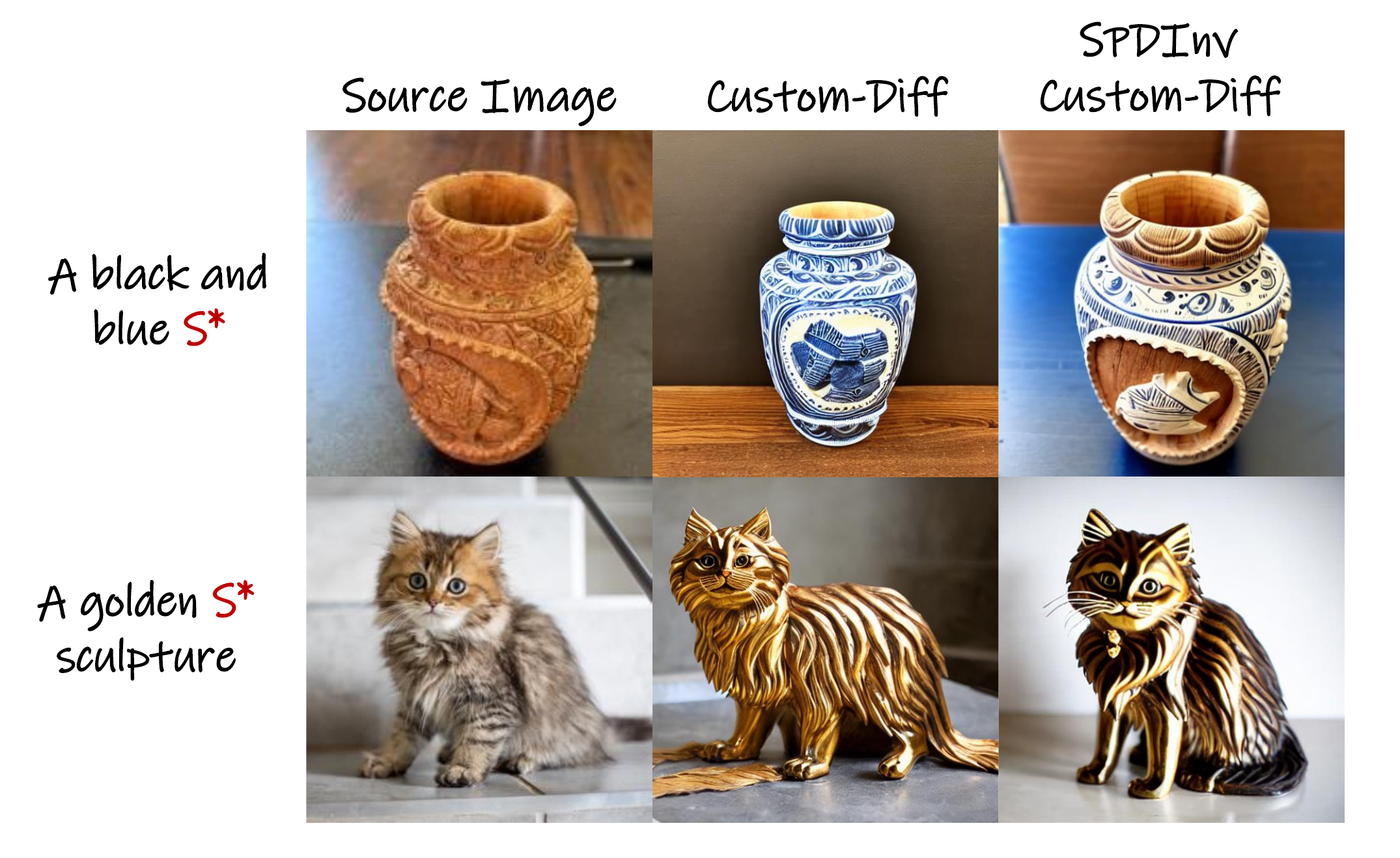}
\vspace{-3mm}
  \caption{Visual results of localized image editing with custom-diff.}
  \label{fig:SP_custom}
\end{figure}

Except for ELITE\cite{wei2023elite}, we integrate our SPDInv into another customized generation method, \eg, Custom-Diff\cite{choi2023custom}. Leveraging its pre-trained model on two distinct concepts, pot (row 1) and cat (row 2), we perform localized editing and present the corresponding visual results in \cref{fig:SP_custom}. One can see that the embedding of SPDInv endows Custom-Diff the ability to maintain consistent layout and pose, while the original Custom-Diff exhibits very different layout and pose from the input images.

\section{Failure Cases}
\begin{figure}[t]
  \centering
  \includegraphics[width=0.95\linewidth]{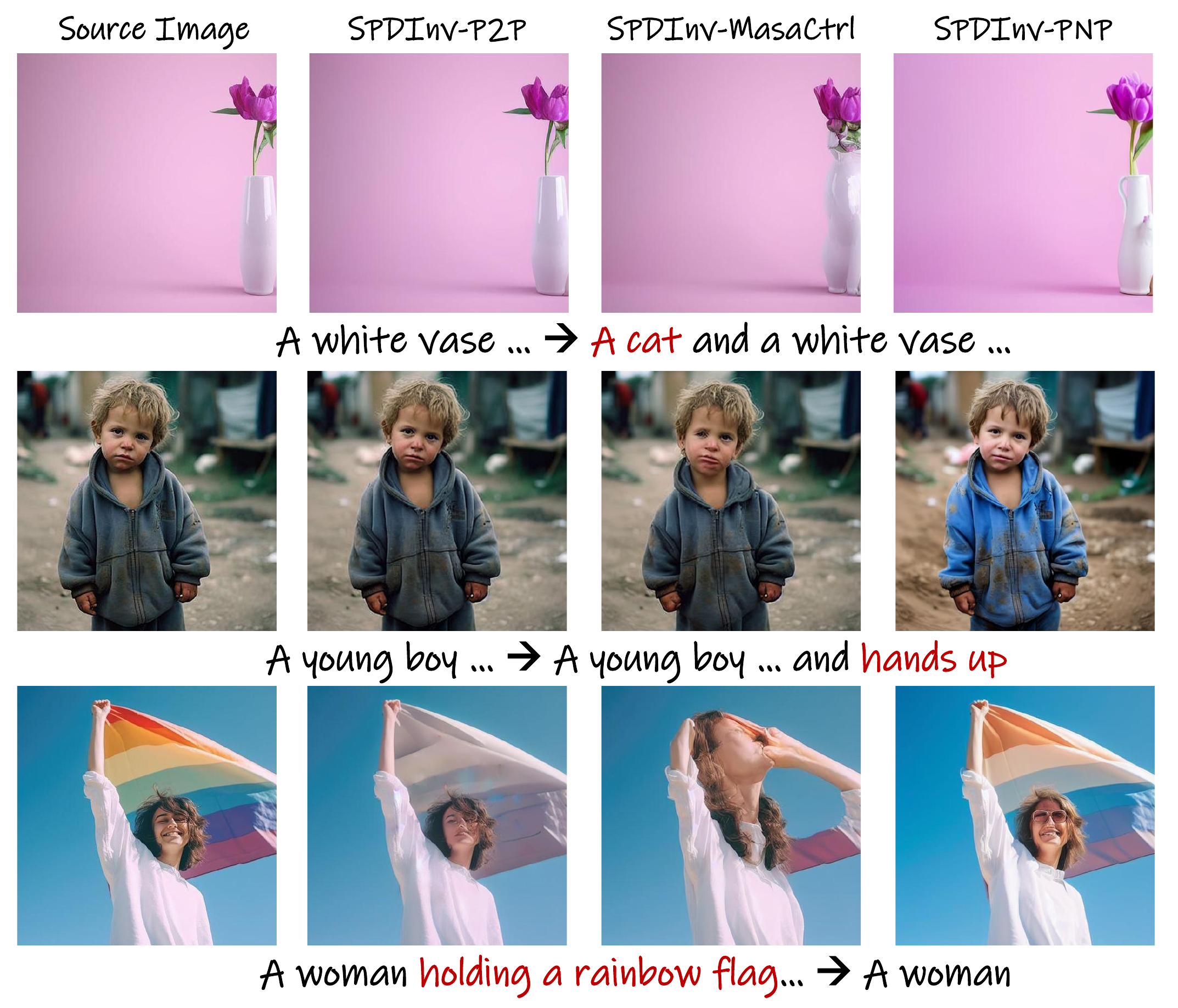}
\vspace{-3mm}
  \caption{Failure cases of SPDInv.}
  \label{fig:SP_failre}
\end{figure}

As we stated in the conclusion of the main paper, although SPDInv achieves great improvement on the overall editing performance, it may fail in cases such as adding, dropping items and editing portrait with current editing engines (\ie, P2P\cite{Prompt2Prompt}, MasaCtrl\cite{masactrl}, PNP\cite{plugandplay}). Some failure cases are depicted in \cref{fig:SP_failre}. Future work will be conducted for improving the stability and robustness of SPDInv on these editing scenarios.

\clearpage

\bibliographystyle{splncs04}
\bibliography{ECCV}

\begin{thebibliography}{10}
\providecommand{\url}[1]{\texttt{#1}}
\providecommand{\urlprefix}{URL }
\providecommand{\doi}[1]{https://doi.org/#1}

\bibitem{avrahami2022blended}
Avrahami, O., Lischinski, D., Fried, O.: Blended diffusion for text-driven editing of natural images. In: Proceedings of the IEEE/CVF Conference on Computer Vision and Pattern Recognition. pp. 18208--18218 (2022)

\bibitem{blattmann2023align}
Blattmann, A., Rombach, R., Ling, H., Dockhorn, T., Kim, S.W., Fidler, S., Kreis, K.: Align your latents: High-resolution video synthesis with latent diffusion models. In: Proceedings of the IEEE/CVF Conference on Computer Vision and Pattern Recognition. pp. 22563--22575 (2023)

\bibitem{brooks2023instructpix2pix}
Brooks, T., Holynski, A., Efros, A.A.: Instructpix2pix: Learning to follow image editing instructions. In: Proceedings of the IEEE/CVF Conference on Computer Vision and Pattern Recognition. pp. 18392--18402 (2023)

\bibitem{masactrl}
Cao, M., Wang, X., Qi, Z., Shan, Y., Qie, X., Zheng, Y.: Masactrl: Tuning-free mutual self-attention control for consistent image synthesis and editing. arXiv preprint arXiv:2304.08465  (2023)

\bibitem{caron2021emerging}
Caron, M., Touvron, H., Misra, I., J{\'e}gou, H., Mairal, J., Bojanowski, P., Joulin, A.: Emerging properties in self-supervised vision transformers. In: Proceedings of the IEEE/CVF international conference on computer vision. pp. 9650--9660 (2021)

\bibitem{chai2023stablevideo}
Chai, W., Guo, X., Wang, G., Lu, Y.: Stablevideo: Text-driven consistency-aware diffusion video editing. In: Proceedings of the IEEE/CVF International Conference on Computer Vision. pp. 23040--23050 (2023)

\bibitem{cho2024noise}
Cho, H., Lee, J., Kim, S.B., Oh, T.H., Jeong, Y.: Noise map guidance: Inversion with spatial context for real image editing. arXiv preprint arXiv:2402.04625  (2024)

\bibitem{choi2023custom}
Choi, J., Choi, Y., Kim, Y., Kim, J., Yoon, S.: Custom-edit: Text-guided image editing with customized diffusion models. arXiv preprint arXiv:2305.15779  (2023)

\bibitem{elarabawy2022direct}
Elarabawy, A., Kamath, H., Denton, S.: Direct inversion: Optimization-free text-driven real image editing with diffusion models. arXiv preprint arXiv:2211.07825  (2022)

\bibitem{epstein2024diffusion}
Epstein, D., Jabri, A., Poole, B., Efros, A., Holynski, A.: Diffusion self-guidance for controllable image generation. Advances in Neural Information Processing Systems  \textbf{36} (2024)

\bibitem{gal2022image}
Gal, R., Alaluf, Y., Atzmon, Y., Patashnik, O., Bermano, A.H., Chechik, G., Cohen-Or, D.: An image is worth one word: Personalizing text-to-image generation using textual inversion. arXiv preprint arXiv:2208.01618  (2022)

\bibitem{guo2023animatediff}
Guo, Y., Yang, C., Rao, A., Wang, Y., Qiao, Y., Lin, D., Dai, B.: Animatediff: Animate your personalized text-to-image diffusion models without specific tuning. arXiv preprint arXiv:2307.04725  (2023)

\bibitem{Prox}
Han, L., Wen, S., Chen, Q., Zhang, Z., Song, K., Ren, M., Gao, R., Chen, Y., Liu, D., Zhangli, Q., et~al.: Improving negative-prompt inversion via proximal guidance. arXiv preprint arXiv:2306.05414  (2023)

\bibitem{Prompt2Prompt}
Hertz, A., Mokady, R., Tenenbaum, J., Aberman, K., Pritch, Y., Cohen-Or, D.: Prompt-to-prompt image editing with cross attention control. arXiv preprint arXiv:2208.01626  (2022)

\bibitem{DDPM}
Ho, J., Jain, A., Abbeel, P.: Denoising diffusion probabilistic models. Advances in neural information processing systems  \textbf{33},  6840--6851 (2020)

\bibitem{CFG}
Ho, J., Salimans, T.: Classifier-free diffusion guidance. arXiv preprint arXiv:2207.12598  (2022)

\bibitem{hu2021lora}
Hu, E.J., Shen, Y., Wallis, P., Allen-Zhu, Z., Li, Y., Wang, S., Wang, L., Chen, W.: Lora: Low-rank adaptation of large language models. arXiv preprint arXiv:2106.09685  (2021)

\bibitem{hu2023animate}
Hu, L., Gao, X., Zhang, P., Sun, K., Zhang, B., Bo, L.: Animate anyone: Consistent and controllable image-to-video synthesis for character animation. arXiv preprint arXiv:2311.17117  (2023)

\bibitem{huang2023reversion}
Huang, Z., Wu, T., Jiang, Y., Chan, K.C., Liu, Z.: Reversion: Diffusion-based relation inversion from images. arXiv preprint arXiv:2303.13495  (2023)

\bibitem{huberman2023edit}
Huberman-Spiegelglas, I., Kulikov, V., Michaeli, T.: An edit friendly ddpm noise space: Inversion and manipulations. arXiv preprint arXiv:2304.06140  (2023)

\bibitem{directinv}
Ju, X., Zeng, A., Bian, Y., Liu, S., Xu, Q.: Direct inversion: Boosting diffusion-based editing with 3 lines of code. arXiv preprint arXiv:2310.01506  (2023)

\bibitem{kawar2023imagic}
Kawar, B., Zada, S., Lang, O., Tov, O., Chang, H., Dekel, T., Mosseri, I., Irani, M.: Imagic: Text-based real image editing with diffusion models. In: Proceedings of the IEEE/CVF Conference on Computer Vision and Pattern Recognition. pp. 6007--6017 (2023)

\bibitem{kumari2023multi}
Kumari, N., Zhang, B., Zhang, R., Shechtman, E., Zhu, J.Y.: Multi-concept customization of text-to-image diffusion. In: Proceedings of the IEEE/CVF Conference on Computer Vision and Pattern Recognition. pp. 1931--1941 (2023)

\bibitem{li2023stylediffusion}
Li, S., van~de Weijer, J., Hu, T., Khan, F.S., Hou, Q., Wang, Y., Yang, J.: Stylediffusion: Prompt-embedding inversion for text-based editing. arXiv preprint arXiv:2303.15649  (2023)

\bibitem{li2023photomaker}
Li, Z., Cao, M., Wang, X., Qi, Z., Cheng, M.M., Shan, Y.: Photomaker: Customizing realistic human photos via stacked id embedding. In: IEEE Conference on Computer Vision and Pattern Recognition (CVPR) (2024)

\bibitem{lin2014microsoft}
Lin, T.Y., Maire, M., Belongie, S., Hays, J., Perona, P., Ramanan, D., Doll{\'a}r, P., Zitnick, C.L.: Microsoft coco: Common objects in context. In: Computer Vision--ECCV 2014: 13th European Conference, Zurich, Switzerland, September 6-12, 2014, Proceedings, Part V 13. pp. 740--755. Springer (2014)

\bibitem{liu2023zero}
Liu, R., Wu, R., Van~Hoorick, B., Tokmakov, P., Zakharov, S., Vondrick, C.: Zero-1-to-3: Zero-shot one image to 3d object. In: Proceedings of the IEEE/CVF International Conference on Computer Vision. pp. 9298--9309 (2023)

\bibitem{liu2023syncdreamer}
Liu, Y., Lin, C., Zeng, Z., Long, X., Liu, L., Komura, T., Wang, W.: Syncdreamer: Generating multiview-consistent images from a single-view image. arXiv preprint arXiv:2309.03453  (2023)

\bibitem{meiri2023fixed}
Meiri, B., Samuel, D., Darshan, N., Chechik, G., Avidan, S., Ben-Ari, R.: Fixed-point inversion for text-to-image diffusion models. arXiv preprint arXiv:2312.12540  (2023)

\bibitem{meng2021sdedit}
Meng, C., He, Y., Song, Y., Song, J., Wu, J., Zhu, J.Y., Ermon, S.: Sdedit: Guided image synthesis and editing with stochastic differential equations. arXiv preprint arXiv:2108.01073  (2021)

\bibitem{NPI}
Miyake, D., Iohara, A., Saito, Y., Tanaka, T.: Negative-prompt inversion: Fast image inversion for editing with text-guided diffusion models. arXiv preprint arXiv:2305.16807  (2023)

\bibitem{NTI}
Mokady, R., Hertz, A., Aberman, K., Pritch, Y., Cohen-Or, D.: Null-text inversion for editing real images using guided diffusion models. In: Proceedings of the IEEE/CVF Conference on Computer Vision and Pattern Recognition. pp. 6038--6047 (2023)

\bibitem{mou2023t2i}
Mou, C., Wang, X., Xie, L., Wu, Y., Zhang, J., Qi, Z., Shan, Y., Qie, X.: T2i-adapter: Learning adapters to dig out more controllable ability for text-to-image diffusion models. arXiv preprint arXiv:2302.08453  (2023)

\bibitem{nichol2021glide}
Nichol, A., Dhariwal, P., Ramesh, A., Shyam, P., Mishkin, P., McGrew, B., Sutskever, I., Chen, M.: Glide: Towards photorealistic image generation and editing with text-guided diffusion models. arXiv preprint arXiv:2112.10741  (2021)

\bibitem{AIDI}
Pan, Z., Gherardi, R., Xie, X., Huang, S.: Effective real image editing with accelerated iterative diffusion inversion. In: Proceedings of the IEEE/CVF International Conference on Computer Vision. pp. 15912--15921 (2023)

\bibitem{zero2023parmar}
Parmar, G., Kumar~Singh, K., Zhang, R., Li, Y., Lu, J., Zhu, J.Y.: Zero-shot image-to-image translation. In: ACM SIGGRAPH 2023 Conference Proceedings. pp. 1--11 (2023)

\bibitem{peebles2023scalable}
Peebles, W., Xie, S.: Scalable diffusion models with transformers. In: Proceedings of the IEEE/CVF International Conference on Computer Vision. pp. 4195--4205 (2023)

\bibitem{podell2023sdxl}
Podell, D., English, Z., Lacey, K., Blattmann, A., Dockhorn, T., M{\"u}ller, J., Penna, J., Rombach, R.: Sdxl: Improving latent diffusion models for high-resolution image synthesis. arXiv preprint arXiv:2307.01952  (2023)

\bibitem{poole2022dreamfusion}
Poole, B., Jain, A., Barron, J.T., Mildenhall, B.: Dreamfusion: Text-to-3d using 2d diffusion. arXiv preprint arXiv:2209.14988  (2022)

\bibitem{radford2021learning}
Radford, A., Kim, J.W., Hallacy, C., Ramesh, A., Goh, G., Agarwal, S., Sastry, G., Askell, A., Mishkin, P., Clark, J., et~al.: Learning transferable visual models from natural language supervision. In: International conference on machine learning. pp. 8748--8763. PMLR (2021)

\bibitem{ramesh2022hierarchical}
Ramesh, A., Dhariwal, P., Nichol, A., Chu, C., Chen, M.: Hierarchical text-conditional image generation with clip latents. arXiv preprint arXiv:2204.06125  \textbf{1}(2), ~3 (2022)

\bibitem{LDM}
Rombach, R., Blattmann, A., Lorenz, D., Esser, P., Ommer, B.: High-resolution image synthesis with latent diffusion models. In: Proceedings of the IEEE/CVF conference on computer vision and pattern recognition. pp. 10684--10695 (2022)

\bibitem{ruiz2023dreambooth}
Ruiz, N., Li, Y., Jampani, V., Pritch, Y., Rubinstein, M., Aberman, K.: Dreambooth: Fine tuning text-to-image diffusion models for subject-driven generation. In: Proceedings of the IEEE/CVF Conference on Computer Vision and Pattern Recognition. pp. 22500--22510 (2023)

\bibitem{saharia2022photorealistic}
Saharia, C., Chan, W., Saxena, S., Li, L., Whang, J., Denton, E.L., Ghasemipour, K., Gontijo~Lopes, R., Karagol~Ayan, B., Salimans, T., et~al.: Photorealistic text-to-image diffusion models with deep language understanding. Advances in Neural Information Processing Systems  \textbf{35},  36479--36494 (2022)

\bibitem{shi2023dragdiffusion}
Shi, Y., Xue, C., Pan, J., Zhang, W., Tan, V.Y., Bai, S.: Dragdiffusion: Harnessing diffusion models for interactive point-based image editing. arXiv preprint arXiv:2306.14435  (2023)

\bibitem{DDIM}
Song, J., Meng, C., Ermon, S.: Denoising diffusion implicit models. In: International Conference on Learning Representations (2020)

\bibitem{song2020score}
Song, Y., Sohl-Dickstein, J., Kingma, D.P., Kumar, A., Ermon, S., Poole, B.: Score-based generative modeling through stochastic differential equations. arXiv preprint arXiv:2011.13456  (2020)

\bibitem{tewel2023key}
Tewel, Y., Gal, R., Chechik, G., Atzmon, Y.: Key-locked rank one editing for text-to-image personalization. In: ACM SIGGRAPH 2023 Conference Proceedings. pp. 1--11 (2023)

\bibitem{tsaban2023ledits}
Tsaban, L., Passos, A.: Ledits: Real image editing with ddpm inversion and semantic guidance. arXiv preprint arXiv:2307.00522  (2023)

\bibitem{plugandplay}
Tumanyan, N., Geyer, M., Bagon, S., Dekel, T.: Plug-and-play diffusion features for text-driven image-to-image translation. In: Proceedings of the IEEE/CVF Conference on Computer Vision and Pattern Recognition. pp. 1921--1930 (2023)

\bibitem{voronov2024loss}
Voronov, A., Khoroshikh, M., Babenko, A., Ryabinin, M.: Is this loss informative? faster text-to-image customization by tracking objective dynamics. Advances in Neural Information Processing Systems  \textbf{36} (2024)

\bibitem{wallace2023edict}
Wallace, B., Gokul, A., Naik, N.: Edict: Exact diffusion inversion via coupled transformations. In: Proceedings of the IEEE/CVF Conference on Computer Vision and Pattern Recognition. pp. 22532--22541 (2023)

\bibitem{wang2024instantid}
Wang, Q., Bai, X., Wang, H., Qin, Z., Chen, A.: Instantid: Zero-shot identity-preserving generation in seconds. arXiv preprint arXiv:2401.07519  (2024)

\bibitem{wang2023stylediffusion}
Wang, Z., Zhao, L., Xing, W.: Stylediffusion: Controllable disentangled style transfer via diffusion models. In: Proceedings of the IEEE/CVF International Conference on Computer Vision. pp. 7677--7689 (2023)

\bibitem{wei2023elite}
Wei, Y., Zhang, Y., Ji, Z., Bai, J., Zhang, L., Zuo, W.: Elite: Encoding visual concepts into textual embeddings for customized text-to-image generation. arXiv preprint arXiv:2302.13848  (2023)

\bibitem{xu2023inversion}
Xu, S., Huang, Y., Pan, J., Ma, Z., Chai, J.: Inversion-free image editing with natural language. arXiv preprint arXiv:2312.04965  (2023)

\bibitem{yang2023paint}
Yang, B., Gu, S., Zhang, B., Zhang, T., Chen, X., Sun, X., Chen, D., Wen, F.: Paint by example: Exemplar-based image editing with diffusion models. In: Proceedings of the IEEE/CVF Conference on Computer Vision and Pattern Recognition. pp. 18381--18391 (2023)

\bibitem{ye2023ip}
Ye, H., Zhang, J., Liu, S., Han, X., Yang, W.: Ip-adapter: Text compatible image prompt adapter for text-to-image diffusion models. arXiv preprint arXiv:2308.06721  (2023)

\bibitem{zhang2023exact}
Zhang, G., Lewis, J.P., Kleijn, W.B.: Exact diffusion inversion via bi-directional integration approximation. arXiv preprint arXiv:2307.10829  (2023)

\bibitem{zhang2023adding}
Zhang, L., Rao, A., Agrawala, M.: Adding conditional control to text-to-image diffusion models. In: Proceedings of the IEEE/CVF International Conference on Computer Vision. pp. 3836--3847 (2023)

\bibitem{zhang2024real}
Zhang, Y., Xing, J., Lo, E., Jia, J.: Real-world image variation by aligning diffusion inversion chain. Advances in Neural Information Processing Systems  \textbf{36} (2024)

\bibitem{zhang2023inversion}
Zhang, Y., Huang, N., Tang, F., Huang, H., Ma, C., Dong, W., Xu, C.: Inversion-based style transfer with diffusion models. In: Proceedings of the IEEE/CVF Conference on Computer Vision and Pattern Recognition. pp. 10146--10156 (2023)

\end{thebibliography}
\end{document}